# BOUNDING BOX-FREE INSTANCE SEGMENTATION USING SEMI-SUPERVISED ITERATIVE LEARNING FOR GENERATING A CITY-SCALE VEHICLE DATASET

Osmar Luiz Ferreira de Carvalho [1], Osmar Abílio de Carvalho Júnior [2, *],
Anesmar Olino de Albuquerque [2], Nickolas Castro Santana [2], Dibio Leandro Borges [1],
Roberto Arnaldo Trancoso Gomes [2], Renato Fontes Guimarães [2]

1 - Departamento de Ciência da Computação, Universidade de Brasília, Campus Universitário Darcy Ribeiro, Asa Norte, Brasília, DF. CEP.: 70910-900

osmarcarvalho@ieee.org and dibio@unb.br

2 - Departamento de Geografia, Universidade de Brasília, Campus Universitário Darcy Ribeiro, Asa Norte, Brasília, DF. CEP.: 70910-900

anesmar@ieee.org, nickolas.santana@unb.br, robertogomes@unb.br, renatofg@unb.br, osmarjr@unb.br

\* Correspondence: osmarjr@unb.br

**ABSTRACT**

Vehicle classification is a hot computer vision topic, with studies ranging from ground-view up to top-view imagery. In remote sensing, the usage of top-view images allows for understanding city patterns, vehicle concentration, traffic management, and others. However, there are some difficulties when aiming for pixel-wise classification: (a) most vehicle classification studies use object detection methods, and most publicly available datasets are designed for this task, (b) creating instance segmentation datasets is laborious, and (c) traditional instance segmentation methods underperform on this task since the objects are small. Thus, the present research objectives are: (1) propose a novel semi-supervised iterative learning approach using GIS software, (2) propose a box-free instance segmentation approach, and (3) provide a city-scale vehicle dataset. The iterative learning procedure considered: (1) label a small number of vehicles, (2) train on those samples, (3) use the model to classify the entire image, (4) convert the image prediction into a polygon shapefile, (5) correct some areas with errors and include them in the training data, and (6) repeat until results are satisfactory. To separate instances, we considered vehicle interior and vehicle borders, and the DL model was the U-net with the Efficient-net-B7 backbone. When removing the borders, the vehicle interior becomes isolated, allowing for unique object identification. To recover the deleted 1-pixel borders, we proposed a simple method to expand each prediction. The results show better pixel-wise metrics when compared to the Mask-RCNN (82% against 67% in IoU). On per-object analysis, the overall accuracy, precision, and recall were greater than 90%. This pipeline applies to any remote sensing target, being very efficient for segmentation and generating datasets.

**1. INTRODUCTION**

Usually, the city's infrastructure was not designed to absorb population growth and road traffic, which has reached high congestion levels in many urban centers worldwide. The accentuated growth in the number of vehicles makes monitoring and managing urban traffic highly complex and necessary. In this context, automatic vehicle detection based on remote sensing images is a powerful tool for various applications such as traffic monitoring, air pollution, congestion studies, public safety, parking utilization, disaster management, and rescue missions. Periodic image acquisition provides information on the number and location of vehicles in different urban environments, allowing coverage of large areas and proper monitoring of moving targets.

Vehicle detection is a widely studied topic in the computer vision community, containing several studies with ground-view and aerial-view images. These two approaches present marked differences in vehicle representation, in which ground images emphasize the vehicle faces, while the top view of the vehicle acquires straight shapes (Ji et al., 2019; Sakhare et al., 2020). Another significant difference is that the vehicle's spatial resolution in aerial images is significantly lower than in terrestrial images.

In-ground view imaging, several literature reviews address advanced driver assistance systems (ADAS) for autonomous vehicles using image processing and vehicle detection from various onboard handling sensors such as radar, monocular camera, and camera binocular (Feng et al., 2021; Janai et al., 2020; H. Wang et al., 2019). In addition, several studies use images from surveillance cameras on roads (Song et al., 2019), on top of buildings (Xi et al., 2019), pedestrian bridges (Fachrie, 2020), among others.

Despite the broad applicability of ground images and videos, vehicle detection from high-resolution aerial and satellite images allows for a synoptic understanding of city patterns, guiding crucial public policies such as urban planning and traffic management. Vehicle detection using aerial view imagery includes different strategies and sensors such as unmanned aerial vehicles (UAV), airplanes, or orbital platforms, which provide data at different heights and resolutions.

Even though skilled professionals may easily distinguish vehicles from different urban features, the rapid and automatic classification is a challenging task since the vehicles: (a) are usually small objects, (2) present a high variability in shape, color and size, (3) appear in different background settings, (4) present different brightness and contrasts among the city, (5) may be crowded (e.g., parking lots), (6) may be occluded by other objects, such as trees and buildings, (6) have many look-alikes in the city. Figure 1 shows six examples of difficult areas to identify the vehicles, where A and B presents shadows, C and D presents a large concentration of vehicles, E presents look-alikes (the tombs are very similar to cars when seen from this angle), and F presents occluded cars by the building roof.

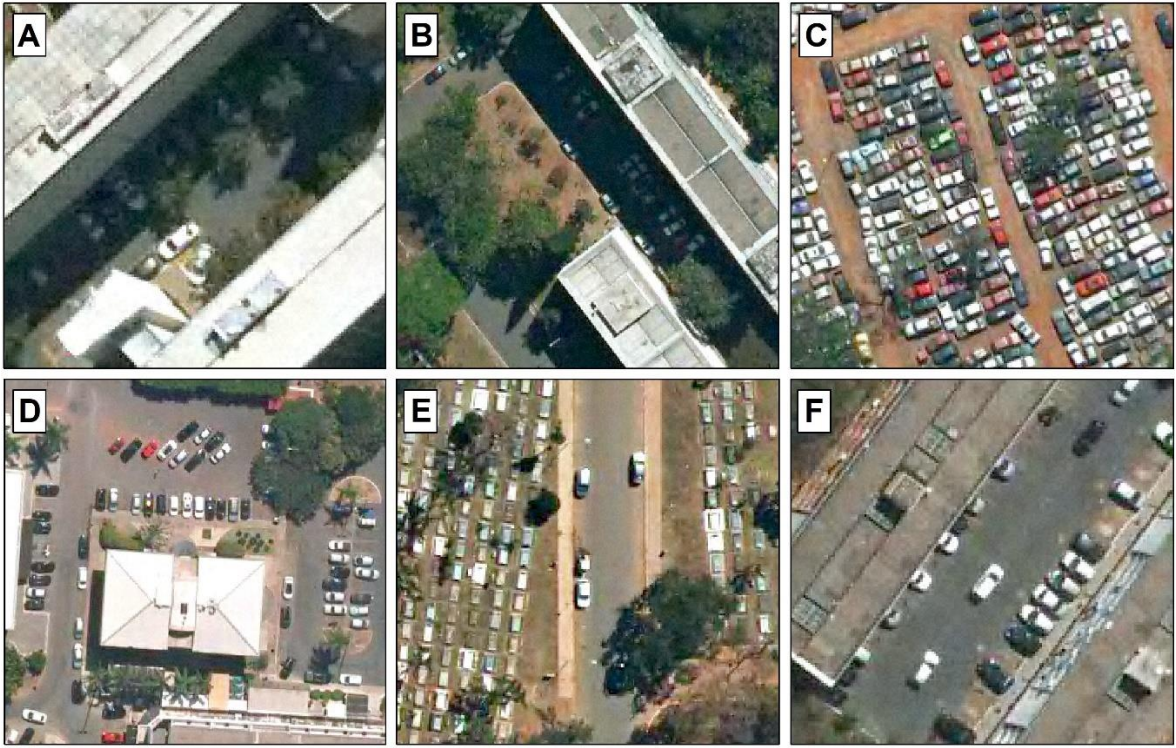

**Figure 1.** Six examples (A, B, C, D, E, and F) of difficult regions to classify cars in the urban setting.

Currently, the Deep Learning (DL) methods represent state-of-the-art vehicle detection, surpassing traditional algorithms. These advances are strongly related to Convolutional Neural Networks (CNN), which apply kernels along with the image, obtaining low, middle, and high-level features, enhancing the classification results. Vehicle detection using deep learning may present different approaches, such as object detection (Zhao et al., 2019), semantic segmentation (Guo et al., 2018), and instance segmentation (Hafiz and Bhat, 2020). In object detection, the DL outputs bounding boxes around the car. Instance segmentation generates bounding boxes and a segmentation mask, and semantic segmentation outputs a class-aware segmentation mask.

Most studies on vehicles address object detection that focuses on the delineation of the targets' bounding box, while instance segmentation, which aims at mapping each object at the pixel level, is still little explored. A challenge in the individual segmentation of vehicles is the lower performance for small objects that, when they are very close, coalesce into a single group (Mou and Zhu, 2018; Tayara et al., 2018). Furthermore, deep instance segmentation methods require a large amount of data, especially considering small object detection. Therefore, training requires a much more complex annotation (since it requires the polygons from each object), containing all possible variations and apparition locations to not depend on a given scenario. The Common Objects in Context (COCO) (Lin et al., 2014) dataset defined small objects with less than $32^2$ pixels, and results considering the small objects are nearly half of the performance on medium and large objects.

More recently, artificial intelligence has an upcoming trend that aims to enhance results and practical solutions by using a data-centric rather than a model-centric approach. The central concept behind this is that the model performance is already very high and that enhancing the data would bring better benefits. One of the pillars for using these methods in selecting the most informative data within a dataset. In this context, active learning is a promising methodology to obtain quality labeled data sequentially. In remote sensing, images often present vast dimensions, and the integration of commonly used GIS software may be an excellent ally for active learning in object detection, since: (a) we may see the entire data at once, (b) It is very straightforward to manipulate and correct polygon data, and (c) we may use other facilities such as polygon shapefiles to choose where to gather the data.

Considering the difficulties and the available tools, the present research aims to advance in three fields (data generation, deep learning method, and dataset):

- **Novel iterative learning procedure for data generation:** a novel iterative learning procedure by integrating DL results with commonly used GIS platforms, which applies to any other target, being less laborious and time-consuming.
- **Bounding Box-Free instance segmentation model:** Proposition of a novel instance segmentation method that uses a semantic segmentation model (in our research the U-net with the Efficient-net-B7 backbone), considering multi-class learning (vehicle interiors and boundaries) to isolate distinct objects into separate instances. Since this method deletes a 1-pixel boundary of each object, we also propose a method to restore the objects' original dimension.
- **BSB Vehicle Dataset:** a city-scale dataset for instance segmentation, semantic segmentation, and object detection with many hard situations (e.g., look-alikes, shadow areas, occlusions) so that other researchers may use or improved the data.

## 2. RELATED WORKS

Different strategies have been developed and described for vehicle detection through aerial and orbital images in the last two decades. In this trajectory, two main approaches stand out (Li et al., 2020; Shen et al., 2021; Shi et al., 2021): (a) methods based on superficial learning and (b) deep learning-based methods.

### 2.1. Early vehicle detection studies using shallow-learning-based approach

Considering vehicle detection approaches based on superficial learning, Hinz, (2003) proposed a generic subdivision into explicit and implicit models. The explicit model describes a vehicle in 2D or 3D (representation of a box or wire-frame structure), considering the car detection from a "top-down" or "bottom-up" model. The implicit model considers the collection of multiple features of a region of the image and their statistics gathered in vectors followed by a classification process (single classifier, combination of classifiers, or hierarchical model). In the present analysis, we considered the following groups of algorithms: (a) Pixel-wise classification and segmentation (including threshold segmentation method, segmentation based on pixel clustering, segmentation based on edge detection and region growth method, segmentation based on inter-frame difference or background difference); (b) object-based classification; object detection (obtaining the bounding box without vehicle segmentation) from multiple features and machine learning within a sliding window approach.

The threshold segmentation method was widely used in different pre-processed images to highlight vehicles, such as Principal Component Analysis (PCA), Bayesian Background Transformation (BBT), and gradient-based method (Sharma et al., 2006); Morphological grayscale method and background difference (vehicle enhancement by subtraction between the original image and the road background image) (Zheng et al., 2013). Cheng et al., (2012) perform pixel-wise classification for vehicle detection using Dynamic Bayesian Networks (DBNs), considering features that comprise pixel-level information and the relationship between neighboring pixels in a region (location analysis of features and color attributes).

Object-based methods use image segmentation to split an image into separated regions and classify them instead of pixels (Hossain and Chen, 2019). Different vehicle detection surveys use object-oriented image classification, considering: (a) eCognition® classification (Holt et al., 2009); (b) segmentation using Otsu Threshold, feature extraction (geometric-shape properties, gray level features, and Hu moments), and statistical classifier (Eikvil et al., 2009); and (c) superpixel-based image segmentation, HOG features and SVM (Chen et al., 2016).

Vehicle detection methods have increased significantly by combining more robust descriptor extraction procedures with machine learning methods for object detection (Table 1). Therefore, vehicle detection uses an image scan through a pre-trained classifier. Among the methods of extraction and selection of features, the most used were: Haar-like features, Histogram of Oriented Gradient (HoG), Histogram of Gabor Coefficient (HGC), and Local Binary Patterns (LBP), Local Steering Kernel (LSK), bag-of-words (BoW) and Scale Invariant Feature Transform (SIFT). Several studies have improved the description of cars by combining different resource extraction methods. The most used machine learning methods were the Support Vector Machines (SVM) and Adaptive Boosting (AdaBoost) in the classification step. However, the literature also describes the use of other methods to compare and improve detection accuracy and efficiency, such as k-Nearest Neighbor (k-NN), Decision Trees (DT), Random Forests (RF), Dynamic Bayesian Network (DBN), Partial Least Squares (PLS). Some associations between feature extraction methods and classifiers had more significant propagation for detecting vehicles such as HoG + SVM (Dalal and Triggs, 2005) and Haar-like + AdaBoost called Viola-Jones (Viola and Jones, 2001). However, the shallow-learning-based methods do not sufficiently describe and generalize for vehicle detection in complex backgrounds. Some studies to minimize errors restricted vehicle detection only along roads, considering the use of masks from a buffer area (Leitloff et al., 2014; Moranduzzo and Melgani, 2014a, 2014b; Nguyen et al., 2007; Zheng et al., 2013), exclusion of objects elevated above a certain height (e.g., buildings and vegetation) from the DEM (Tuermer et al., 2013), and correlation of cars in consecutive frames (Cao et al., 2011). Also, most of these methods are sensitive to the in-plane rotation of objects (detecting only in a specific orientation) and to changes in lighting such as Viola-Jones.

**Table 1.** Studies developed for the detection of cars using different feature extraction approaches (shallow-learning-based features) and classification, in which the feature extraction methods described are: Color Probability Maps (CPM), Haar-like features (Hlf), Histogram of Gabor Coefficients (HGC), Histogram of Oriented Gradients (HoG), Local Binary Patterns (LBP), Local Steering Kernel (LSK), Local Ternary Pattern (LTP), Opponent Histogram (OH), Scale Invariant Feature Transform (SIFT), and Integral Channel features (ICFs). The classification methods are: Adaptive Boosting (AdaBoost), Decision Trees (DT), Deformable Part Model (DPM), Dynamic Bayesian network (DBN), k-Nearest Neighbor (k-NN), Partial Least Squares (PLS), Random Forests (RF), and Support Vector Machines (SVM). The images used in the articles are: Unmanned Aerial Vehicle (UAV), Google Earth (GE), and Wide Area Motion Imagery (WAMI).

| Article | Features | Classifier | Image |
|---|---|---|---|
| (Leberl et al., 2007) | HoG, Hlf, LBP | AdaBoost | aerial |
| (Nguyen et al., 2007) | HoG, Hlf, LBP | AdaBoost | aerial |
| (Grabner et al., 2008) | LBP, HoG, and Hlf | AdaBoost | aerial |
| (Cao et al., 2011) | HoG | SVM | UAV |
| (Gleason et al., 2011) | HoG and HGH | k-NN, SVM, DT, and RF | aerial |
| (Kembhavi et al., 2011) | HoG, CPM, and pairs of pixel comparisons | PLS | aerial |
| (Liang et al., 2012) | HoG and Hlf | AdaBoost and SVM | WAMI |
| (Shao et al., 2012) | HoG, LBP, and OH | SVM | aerial |
| (Tuermer et al., 2013) | HoG | AdaBoost | aerial |
| (Moranduzzo and Melgani, 2014a) | SIFT | SVM | UAV |
| (Moranduzzo and Melgani, 2014b) | HoG | SVM | UAV |
| (Leitloff et al., 2014) | Hlf | AdaBoost and SVM | aerial |
| (Liu and Mattyus, 2015) | ICFs + HoG | AdaBoost | UAV and GE |
| (Madhogaria et al., 2015) | HoG | SVM and Causal MRF | UAV |
| (Razakarivony and Jurie, 2016) | HOG, LBP, and LTP | SVM, DPM, template matching, and Hough Forest | aerial |
| (Xu et al., 2016) | HoG and Hlf | SVM and AdaBoost | UAV |
| (Cao et al., 2017) | SIFT | Multi-Instance Learning | satellite |
| (Xu et al., 2017b) | Hlf + Road Orientation Adjustment | AdaBoost | UAV |
| (Zhou et al., 2018) | LSK + bag-of-words (BoW) | SVM | UAV and satellite |
| (Liu et al., 2019) | LSK + vector of locally aggregated descriptors (VLAD) | Directed-Acyclic-Graph SVM | aerial |

In the transition from traditional to DL methods, some studies use deep architecture only to extract highly descriptive features combined with a machine learning classifier. In this approach, the following propositions stand out: Deep Boltzmann Machines (DBMs) and weakly supervised learning (Han et al., 2015), multilayer deep resource generation model using DBMs and Multiscale Hough Forest Model (Yu et al., 2016, 2015), CNN and Exemplar-SVMs (Cao et al., 2016), and CNN and SVM (Ammour et al., 2017).

**2.2. Deep-learning-based vehicle detection**

A significant milestone in CNN's dominance in computer vision was its success in the ImageNet Large Scale Visual Recognition Challenge in 2012 (Krizhevsky et al., 2017). DL-based vehicle detection studies have intensified in the following years, with an annual increase making it the dominant method today. Deep learning architecture networks perform better than shallow learning-based methods due to the following reasons (Sevo and Avramovic, 2016): (a) operates both for feature extraction and classification; (b) CNN improves automatic feature generation with the ability to learn local characteristics of different orders, inherently exploiting spatial dependence; and (c) less time consuming. Different deep learning approaches have been applied in vehicle detection, such as object detection, semantic segmentation, instance segmentation.

*2.2.1 Object Detection*

Vehicle studies using object detection are dominant due to fast target detection, improving real-time monitoring efficiency. However, these methods do not allow a precise mapping of their contours obtained with semantic and instance segmentation. Table 2 presents the main studies of vehicles using object detection methods. A subdivision of the object detection algorithms are two-stage object detection and one-stage object detection.

Two-step methods first generate several bounding boxes around potential objects called region proposals, and then a classifier determines the object presence. The classification for each potential object slows down the process, focusing on the detection accuracy. As examples of two-stage object detection algorithms highlight Regions with CNN features (R-CNN) (Girshick et al., 2014), its variants Fast R-CNN (Girshick, 2015), Faster R-CNN (Ren et al., 2017), and Mask R-CNN (He et al., 2020).

One-stage object detection processes images through a single neural network, detecting and classifying multiple objects at the same time and ensuring speed. These methods focus on the detection speed but have limitation to detect crowded groups of small objects. Among these algorithms, You Only Look Once (YOLO) (Redmon et al., 2016), You Only Look Twice (YOLT) (Van Etten, 2018) and Single Shot Multibox Detector (SSD) (Liu et al., 2016) are the most prevalent.

**Table 2.** Related works using object detection algorithms, considering the method and data type. The data types are separated in seven categories: (1) satellite, (2) aerial, (3) UAV, (4) ultrahigh-resolution UAV, (5) Google Earth (GE), (6) Cameras top of building, and (7) several. Acronyms for the methods: Residual Feature Aggregation (RFA), Generative Adversarial Network (GAN), and You Only Look Once (YOLO).

| Paper | Method | Data |
|---|---|---|
| (Chen et al., 2014) | Hybrid Deep Convolutional Neural Network (HDNN) | 5 |
| (Qu et al., 2016) | Detection model with two step: (1) proposed Binary Normed Gradients (BING) to extract region proposals, and (2) feature extraction and classification using CNNs. | 1 |
| (Deng et al., 2017) | Detection model based on two CNNs: (1) accurate vehicle proposal network (AVPN) to predict bounding boxes of vehicle-like targets, and (2) vehicle attributes learning network (VALN) for inferring each vehicle's type and orientation. | 2 |
| (Tang et al., 2017) | An improved vehicle detection method based on Faster R-CNN. | 3 |
| (Xu et al., 2017a) | Vehicle detection using the Faster R-CNN | 3 |
| (Zhong et al., 2017) | Method based on Cascaded Convolutional Neural Networks | |
| (Koga et al., 2018) | Method proposition using Hard Example Mining (HEM) to the Stochastic Gradient Descent training of a CNN vehicle classifier. | 7 |
| (Liu et al., 2018) | Real-Time Ground Vehicle Detection based on CNN. | 3 |
| (Zhu et al., 2018) | Development of the Deep Vehicle Counting Framework based on Enhanced-SSD | 4 |
| (Benjdira et al., 2019) | Comparison between YOLOv3 (best model) and Faster R-CNN | 3 |
| (Chen et al., 2019) | Detection model based on two CNNs that adopt the VGG-16 model | 2 |
| (Gao et al., 2019) | EOVNet (Earth observation image-based vehicle detection network), a modified Faster R-CNN. | 7 |
| (Ji et al., 2019) | Improved Faster R-CNN with Multiscale Feature Fusion and Homography Augmentation | 7 |
| (Li et al., 2019) | R3-Net a deep network for multi-oriented vehicle detection | 2 |
| (Shen et al., 2019) | Detection algorithm based on Faster R-CNN | 2 |
| (Sommer et al., 2019) | Systematic investigation of the Fast R-CNN and Faster R-CNN in vehicle detection | 2 |
| (J. Wang et al., 2019) | Detection using YOLOv3 and vehicle tracking using deep appearance features for vehicle re-identification and Kalman filtering for motion estimation | 3 |
| (Xi et al., 2019) | Model based on multi-task cost-sensitive-convolutional neural network (MTCS-CNN) | 6 |
| (Yang et al., 2019) | Novel double focal loss convolutional neural network (DFLCNN) | |
| (Zhang and Zhu, 2019) | Improved YOLOv3 using a sloping bounding box attached to the angle of the target vehicles | 2 |
| (Guo et al., 2020) | Orientation-aware feature fusion single-stage detection (OAFF-SSD) | 3 |
| (Li et al., 2020) | Detection model for different scales based on three steps: (1) image cropping strategy to divide the image into two patches; (2) combination of the original image and two patches into a batch and detect vehicles with a CNN, and (3) proposition of an Outlier-Aware Non-Maximum Suppression. | 3 |
| (Ham et al., 2020) | Comparison among faster R-CNN, R-FCN, and SSD (Best model) | 3 |
| (Jiang et al., 2020) | Optimized Deep Neural Network considering the following steps: feature extraction from Basenet (VGG and ResNet pretrained by ImageNet), object detection combining double shot with misplaced localization strategy, and non-maximum suppression. | 7 |
| (Mandal et al., 2020) | Small-Sized Vehicle Detection Network (AVDNet) (one-stage vehicle detection network) | 2 |
| (Ophoff et al., 2020) | Comparison among four different single-shot object detection networks: D-YOLO (best model), YOLOV2, YOLOV3, and YOLT | 1 |
| (Stuparu et al., 2020) | Vehicle detection based on RetinaNet architecture | 1 |
| (Tan et al., 2020) | Model based on Alexnet network (classification) and Faster R-CNN (target detection) | 1 |
| (Wang and Gu, 2020) | Faster R-CNN with a improved feature-balanced pyramid network (FBPN) | 2 |
| (Ammar et al., 2021) | Comparison among YOLOv3, YOLOv4 (best models), and Faster R-CNN | 3 |
| (Bashir and Wang, 2021) | Super-resolution cyclic GAN with RFA and YOLO as the detection network (SRCGAN-RFA-YOLO) | 1, 2 |
| (Javadi et al., 2021) | Modified YOLOv3 and fcNN using 3D features in cascade. | 2 |
| (Shen et al., 2021) | Method using the lightweight feature extraction network with the Faster R-CNN | 2 |
| (Shi et al., 2021) | Orientation-Aware Vehicle Detection with an Anchor-Free Object Detection approach | 2 |

*2.2.1 Semantic and Instance Segmentation*

Vehicle studies with semantic and instance segmentation present less quantity than those developed with object detection methods. Tayara et al. (2018) performed a Fully Convolutional Regression Network (FCRN), whose training stage uses the input image and ground truth data that describes each vehicle as a 2-D Gaussian function distribution. Therefore, the vehicle's original format acquires a simplified elliptical shape in the ground truth and output images. The vehicle segmentation uses a threshold value in the predicted density map, generating a binary mask. Although the method avoids grouping cars and favors counting, vehicles take on a different form described by the Gaussian function, which has a low precision at the pixel level. In contrast, Mou and Zhu, (2018) sought an instance segmentation of vehicles with pixel-level accuracy, where cars appear well delimited in a distinct physical instance. In this context, a severe problem is the differentiation of vehicles in contact that agglutinated in a single instance. The solution proposed by the authors was to establish an architecture that subdivided the central vehicle regions and their limits instead of treating the vehicle problem as a single unit. Recently, Reksten and Salberg, (2021) used the Mask R-CNN with an image normalization strategy to suit different environments and an accurate road mask to filter driving vehicles from those parked.

Other studies combine a prior segmentation followed by a vehicle detection. Audebert et al., (2017) used the deep-learning-based segment-before-detect method containing three steps: (a) semantic segmentation using a fully convolutional network to infer pixel-level class masks; (b) vehicle detection by regressing the bounding boxes of connected components; and (c) object-level classification using CNN architectures (LeNet, AlexNet, and VGG-16). Yu et al., (2019) developed a convolutional capsule network with the following steps: (a) superpixel segmented, (2) labeling patches into vehicles or background using convolutional capsule network, and (3) non-maximum suppression to eliminate repetitive detections. Tao et al., (2019) performed a scene classification with deep learning followed by different vehicle detectors and post-processing rules according to the scene context.

**3. MATERIAL AND METHODS**

The present research had the following methodological steps (Figure 1): (2.1) dataset creation; (2.2) DL models; and (2.3) model evaluation.

## 3.1 Study Area & Image Acquisition

The entire city of Brasilia was the study area (Figure 2). The use of large regions with many mapped look-alike features and different scenarios favors learning DL models. The Brasilia image has 57,856 x 42,496 spatial dimensions and 0.24-meter resolution obtained by the Infraestrutura de Dados Espaciais do Distrito Federal (IDE/DF) (https://www.geoportal.seduh.df.gov.br/geoportal/, accessed on September 15, 2021). In this scenario, a car has approximately 20 (length) x 10 (width) pixel dimensions.

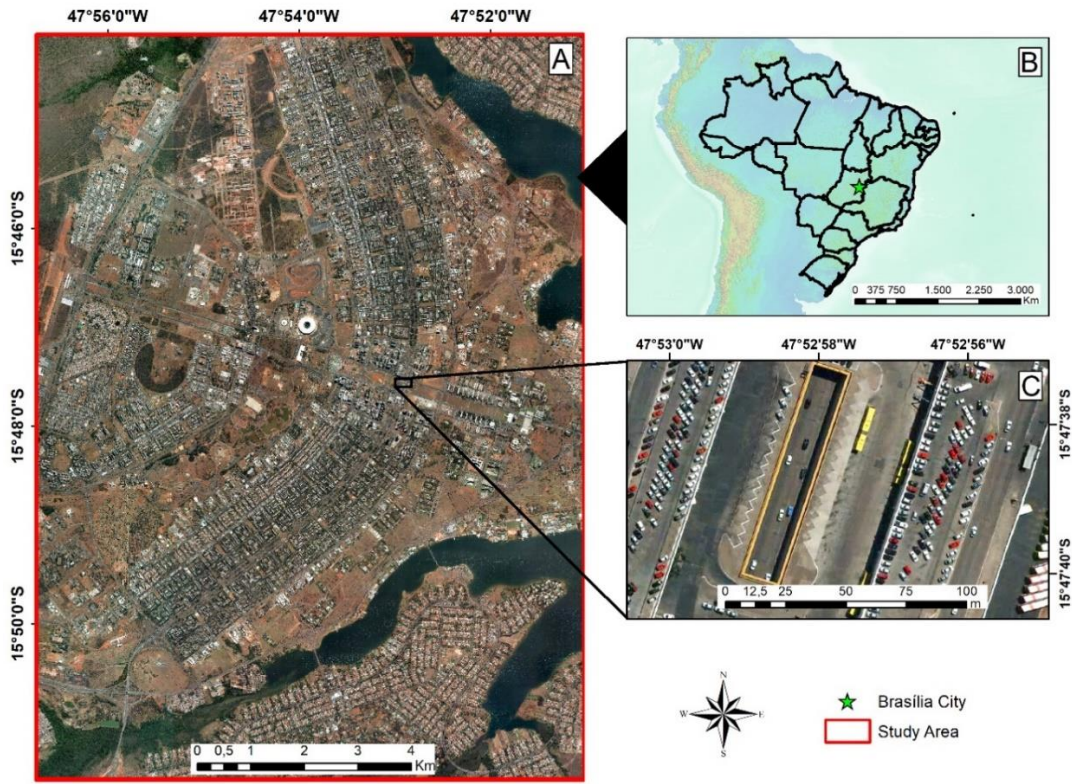

**Figure 2.** Study area.

## 3.2 Semi-supervised Iterative Approach

Manually identifying all the cars in a city is a very time-consuming process. So, the solution is to seek alternatives to automate the generation of datasets correctly. For example, if a very good annotator took five seconds to label a single car, it would take over 200 hours to label 150,000 vehicles. Thus, we proposed a novel semi-supervised approach using Geographic Information System (GIS) data to increase operability (Figure 3). Briefly the method consists in labelling a portion of the entire image for training the model, and then use the model to classify the entire 57,856 x 42,496-pixel image. We convert the predictions into the shapefile format which is straightforward for editing in ArcMap, and correct the areas that present most errors, and include them in the training data.

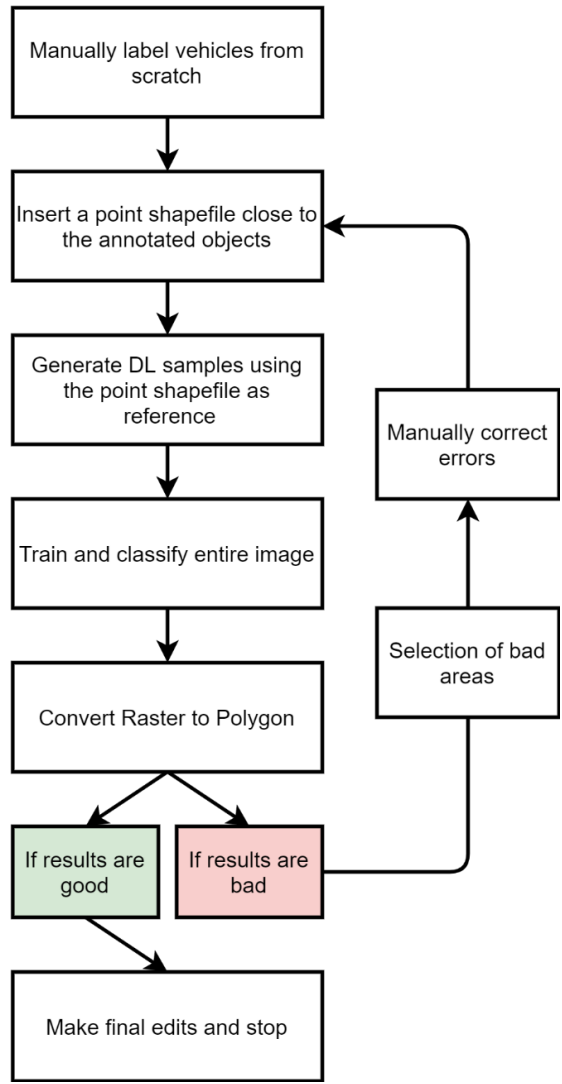

**Figure 3.** Proposed semi-supervised pipeline. (**EDITORIAL INFO: 1 column figure**)

The proposed procedure to increase the training database reconciles an incremental and cumulative learning, selecting samples that improve model performance. An effective database expansion design aims to achieve greater incremental accuracy in subsequent predictions. The procedure is cumulative, using the entire set of labeled samples present in each step. Thus, the segmentation model increases its performance until the accuracy values do not vary significantly, i.e., the decrease in the incremental accuracy is due to the depletion of informative data.

*3.2.1 Ground Truth Using ArcMap*

The manual annotations and corrections used the ArcMap software considering a polygon shapefile for each vehicle, since it is much easier to manipulate when compared to raster (mask) data. We applied a 1-pixel buffer (0.24 meter in the corresponding image) with negative distance to generate the borders inside the polygon features. For the first training procedure, all training samples were made from scratch. For the subsequent iterations, we used the DL predictions as a primary raw data, in which we corrected areas with more errors. The number of verified and corrected areas increase after each iteration using the semi-supervised approach, increasing the dataset.

*3.2.2 Deep Learning Sample Generator software*

To capture the exact corrected areas, it is crucial to generate the DL samples in the strategic areas. Thus, we proposed a novel method for selecting samples using the Point shapefile. This procedure allows choosing critical points where wrong predictions becomes part of new training after correction, quickly improving the model's detection capacity with much less laborious work.

The developed DL sample generator from Point Shapefiles became a module in the Abilius Software program that receives three inputs: (1) the original image; (2) the ground truth image; and (3) the point shapefiles. The program requires inputs in the same projection, and the user may choose the size of the image tiles generated. The software uses the point shapefile to center the image tiles and crops the image and its corresponding ground truth image. Besides, this software outputs the annotations for instance segmentation, considering the COCO annotation format (Lin et al., 2014), which is compatible with Region CNN methods (Girshick et al., 2016), such as the Mask-RCNN (He et al., 2020) and similar methods. Using point shapefiles also enables the user to generate samples close to each other, a powerful augmentation technique.

*3.2.3 Deep Learning Approach*

Usually, region-based instance segmentation underperforms on small objects, and semantic segmentation does not present distinct classification for different instances, unable to differentiate adjacent vehicles. The conversion of a conventional semantic segmentation model to a polygon shapefile with touching vehicles (Figure 4A) are represented by a single polygon. Semantic segmentation models are the most used among the remote sensing community, mainly because of the good per-pixel results and simplicity of models and annotation formats. Thus, to solve this problem, we adopted a similar solution proposed by Mou and Zhu (2018). Instead of multitasking learning, we adopted a multiclass learning procedure, in which the contour class competes against the vehicle class.

The model output subdivides the vehicle into two parts (edge and interior) (Figure 4B). Deleting the edges isolates the individual vehicles, and all previously touching cars will be at least 2 pixels apart from each other. The next step is to develop a function to attribute a different value to each vehicle. This proposed method generates a list with all contours (using the find_contours function from OpenCV (Bradski and Kaehler, 2008)) and iteratively converts the contours to a mask attributing different values from 1 to N, being N the total number of distinct vehicles (Figure 4C). Aiming to optimize computational resources, we adapted the polygon2mask function from scikit-image package (van der Walt et al., 2014) that generates an array with zeros every time it is called, which is costly due to the enormous image dimensions. Thus, we only create an array with zeros once. For each iteration, we attribute different values to the generated mask (one object at a time), guaranteeing distinct values for each vehicle.

Now, the predictions are distinct for each object. However, since the objects are small, a 1-pixel error at the edges is considerable and not as precise. The edge restoration uses the instance array as the input. The first step is to apply 1-pixel padding in the entire image. Then we make eight copies of the original array dislocated in different directions: (1) up, (2) down, (3) left, (4) right, (5) up-right, (6) up-left, (7) down-right, and (8) down-left. Then we sum all arrays considering only pixels with zero value, and remove the initial padding (recovering the image's original shape). This procedure enlarges the object edges, independent of the object orientation, resulting in the same semantic information (Figure 4A), but with different instances for each object (Figure 4D).

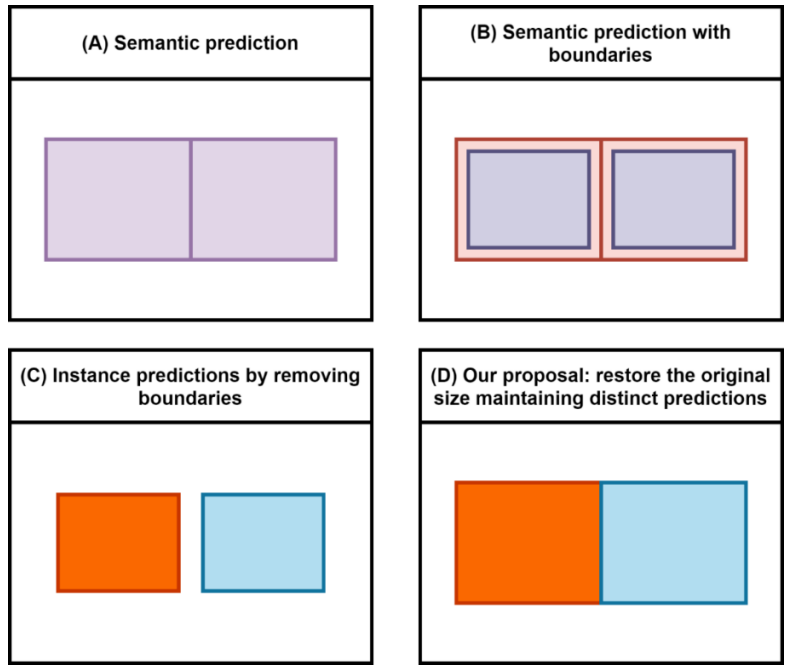

**Figure 4.** Theoretical outputs from semantic segmentation algorithms, in which A is a normal semantic segmentation strategy, B is segmentation with boundaries, C is instance segmentation by removing the boundaries, and D is our proposed solution to restore the correct size maintaining distinct predictions.

Despite the wide variety of semantic segmentation models, we used a single combination throughout the entire process since our primary goal is not to develop a new DL architecture but to make an efficient procedure for large areas per-pixel vehicle detection separating different instances. The configuration used the Semantic Segmentation Models repository (Yakubovskiy, 2020) and considered the U-net architecture (Ronneberger et al., 2015) with the Efficient-net-B7 backbone (Tan and Le, 2019). The U-net is one of the most used models for semantic segmentation purposes, consisting of an encoder-decoder architecture. The encoder is responsible for extracting features, and the decoder is responsible for recovering the image's original dimension with the correct pixel locations. The feature extraction section has much potential to grow with upcoming new methods, and recently the Efficient-net-B7 presented state-of-the-art results.

Maintaining the same hyperparameters ensured consistency between the different phases of training: (a) 300 epochs, (b) Adam optimizer, (c) batch size of five. Besides, the method considered the cross-entropy loss function with weights (0.1 for background, 0.6 for vehicles, 0.3 for the contour) and 15% of the images as validation, saving the model with the lowest cross-entropy loss. The dataset expansion used two augmentation strategies: the random horizontal and vertical flip, both with probabilities of 50%.

Moreover, we compared the proposed method with the Mask-RCNN model (He et al., 2020) using the ResNeXt-101 (Xie et al., 2017) backbone (the best configuration for the Detectron2 (Wu et al., 2019) considering instance segmentation). The two analyzed scenarios were with: (a) image scaling augmentation (transforming the image dimensions from 256x256 up to 800x800) and, (b) without any scaling. In both cases, the augmentation strategies considered the random horizontal and vertical flip, just like the semantic segmentation model. The training used 5,000 iterations, and two images per batch, and the other parameters as default.

*3.2.4 Classifying the Entire Image*

The training images dimensions are 256x256, which is smaller than the entire image dimensions. Thus, to classify the whole image, we considered a sliding window approach with a 128-pixel stride (half the image dimensions). Using a stride smaller than the image dimensions results in overlapping pixels. A traditional way is to take the mean average among the overlapping pixels. Moreover, this approach reduces errors at the borders of the frames, exemplified in recent works (Costa et al., 2021; da Costa et al., 2021; de Albuquerque et al., 2020). A drawback of using this method is the computational cost. The time to classify an image increases by 4 when dividing the stride value by two. Since our image presents large dimensions, we did not consider smaller stride values.

**3.3 Model evaluation**

*3.3.2 Testing Set and Test Areas*

For evaluating the models, we considered a test set of 50 images with 256x256-pixel dimensions (same as dimensions for training and validation), and three independent testing areas (Figure 5), considering different difficulty scenarios. The first considered areas with no occlusion and significant difficulties for the cars (Figure 5A), with 2560x2560-pixel dimensions. The second scenario is a parking lot with many crowded vehicles (Figure 5B) with 2304x2304-pixel dimensions. The third scenario cover residential areas with a building generating shadow and regions of occlusion (Figure 5C) with 1560x1560-pixel dimensions.

The semantic segmentation of the entire test area used a sliding window with 128-pixel steps. Meanwhile, the instance segmentation (Mask-RCNN) of the testing areas used the mosaic method developed by Carvalho et al. (2021).

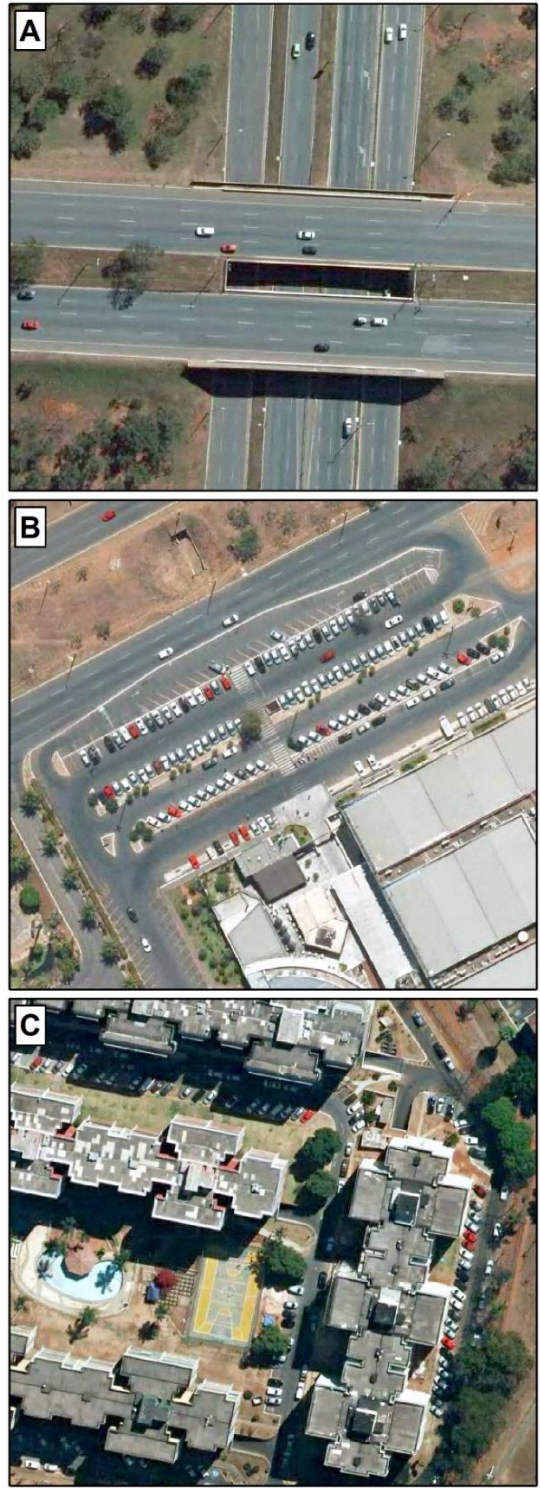

**Figure 5.** Zoom from the three separate testing areas A, B, and C.

*3.3.1 Metrics*

In supervised learning tasks, the accuracy analysis compares the predicted results and the ground truth data. The confusion matrix is a standard structure for all tasks, yielding four possible outcomes: true positives (TP), true negatives (TN), false positives (FP), and false negatives (FN). For semantic segmentation tasks, the confusion matrix analysis is per pixel. There are many possible metrics such as overall accuracy, precision, recall, f-score, among others. However, the intersection over union (IoU) is the primary metric for many works, given by:

$$IoU = \frac{|A \cap B|}{|A \cup B|} = \frac{TP}{TP + FP + FN}$$

In which:

A ∩ B: the area of intersection;

A ∪ B: the area of union.

The accuracy analysis considered: (a) the pixel-wise IoU for the test set and the three testing areas for our proposed model (considering the proposed expanded border algorithm and without considering the borders) and the Mask-RCNN, and (b) the per-object analysis in the test areas for our proposed method. The object analysis had four classifications: (a) correct predictions, (b) partial predictions, (c) false positives, and (d) false negatives.

**4. RESULTS**

**4.1 Training Iterations**

The final version of the dataset used a total of five iterations. The total number of point shapefiles was 1066 with training samples in various scenarios (Figure 6). For each iteration, we considered point shapefiles in areas that the errors did not disappear in previous iterations (to see if the mistakes disappeared). Still, at each iteration, the concentration of points had different focuses. For example, the second training focused on eliminating look-alike features, which already gives a good boost in performance metrics, with an easy correct the error, since we only need to delete some polygons. The fourth training had the minimum number of points since the areas required more corrections (e.g., parking lots), being more laborious. Thus, the proposed procedure effectively uses the results of the DL model in repeated corrections of pseudo-labels. Gradually, the predictions become more reliable, minimizing errors and manual correction labor in each interaction.

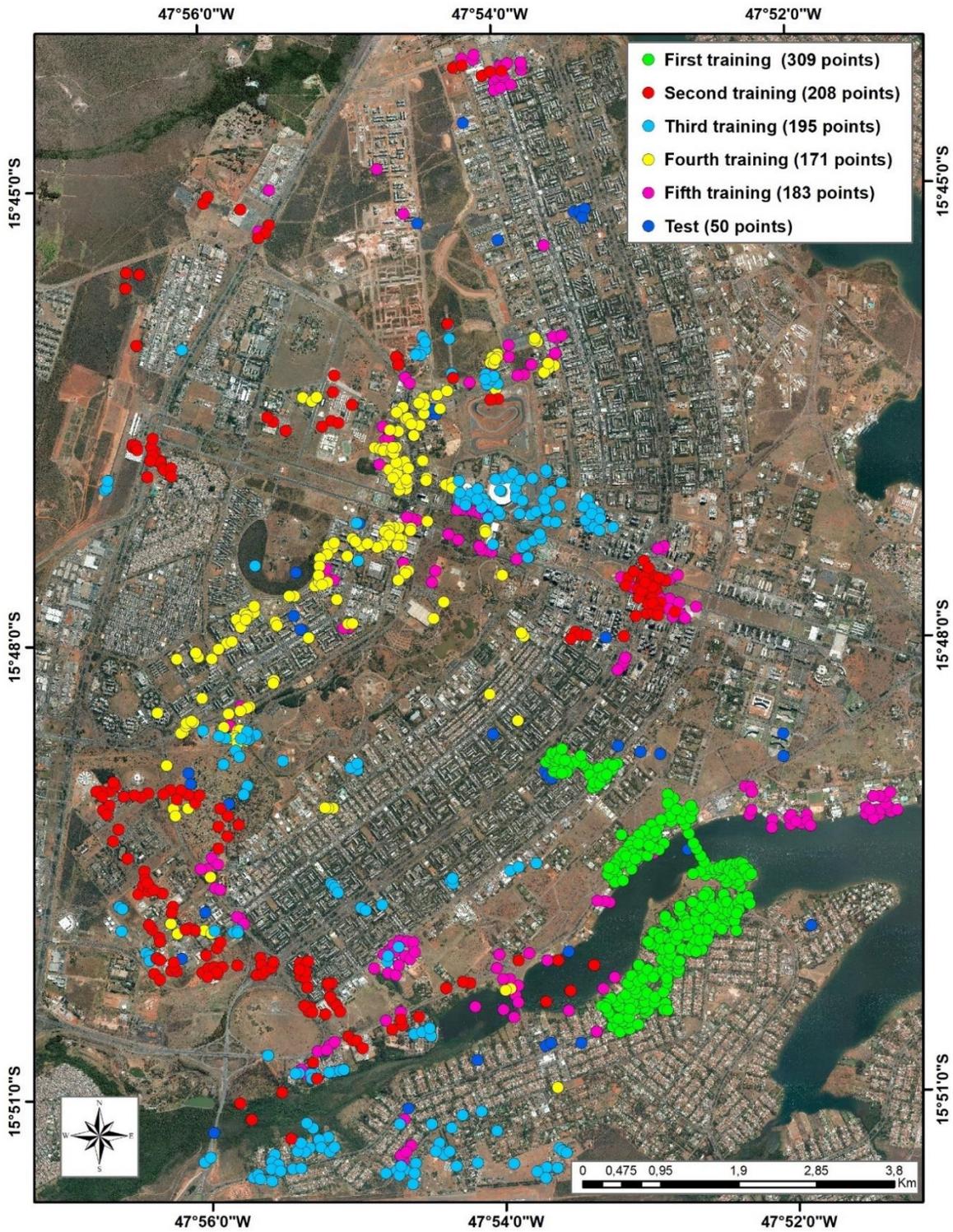

**Figure 6.** Study area with the Point Shapefiles (training points) used in each training, in which the training is cumulative.

## 4.2 Metrics

*4.2.1 Pixel Metrics*

Table 3 lists the results for IoU on the four separate testing sets (Test Area 1, Test Area 2, Test Area 3, and Test Set), considering each training step. There is an evident rise in the metrics when increasing the number of training samples on the same independent test areas. Test Area 1 had the highest results, and it is indeed the easiest since there are no shadows and occluded cars. Test area 2 has a parking lot, with many crowded vehicles, and other vehicles, presenting more errors. Test Area 3 has many regions with shadows, and partial vehicles had the lowest IoU, bringing to light the difficulty in some areas, even for human specialists. The test set has fifty 256x256 samples all around the city, with varying levels of difficulty. The IoU of the test set is approximately the average of the distinct testing areas (81.88).

When comparing the IoU using our growing border algorithm to recover initial values without considering the borders, the results are very distinct, with a greater than 15% difference in the IoU metric. Also, the metrics remain very similar even when increasing the number of training samples. A possible explanation is error compensation, not bringing insightful information on the testing data.

**Table 3.** IoU results for our proposed method in the BSB vehicle dataset considering the expanded (exp.) border algorithm, and not considering the borders, for each train iteration.

| Train # | Type | Test Area 1 | Test Area 2 | Test Area 3 | Test Set |
|---|---|---|---|---|---|
| 1 | No border | 63.19 | 63.67 | 51.97 | 52.60 |
| 1 | Exp. Border | 80.80 | 77.23 | 66.89 | 66.03 |
| 2 | No border | 64.41 | 64.65 | 54.41 | 63.52 |
| 2 | Exp. Border | 86.73 | 79.94 | 74.75 | 80.39 |
| 3 | No border | 60.40 | 62.27 | 52.43 | 61.49 |
| 3 | Exp. Border | 87.69 | 82.31 | 75.95 | 80.73 |
| 4 | No border | 62.83 | 62.39 | 55.81 | 63.39 |
| 4 | Exp. Border | 88.03 | 81.98 | 78.44 | 81.06 |
| 5 | No border | 63.98 | 64.13 | 56.24 | 64.51 |
| 5 | Exp. Border | 88.37 | 81.31 | 77.10 | 82.45 |

Figure 7 shows the semantic segmentation result, with and without borders. The visual results demonstrate that the proposed method expands vectorially 1 pixel on the edges, consisting of a fast process. Furthermore, the instances show an efficient separation. Figure 7B (second row) demonstrates that not using the borders to differentiate objects, the prediction merges vehicles into a single polygon. Expanding edges on different instances retrieves the same semantic prediction information but with the distinction of the vehicles.

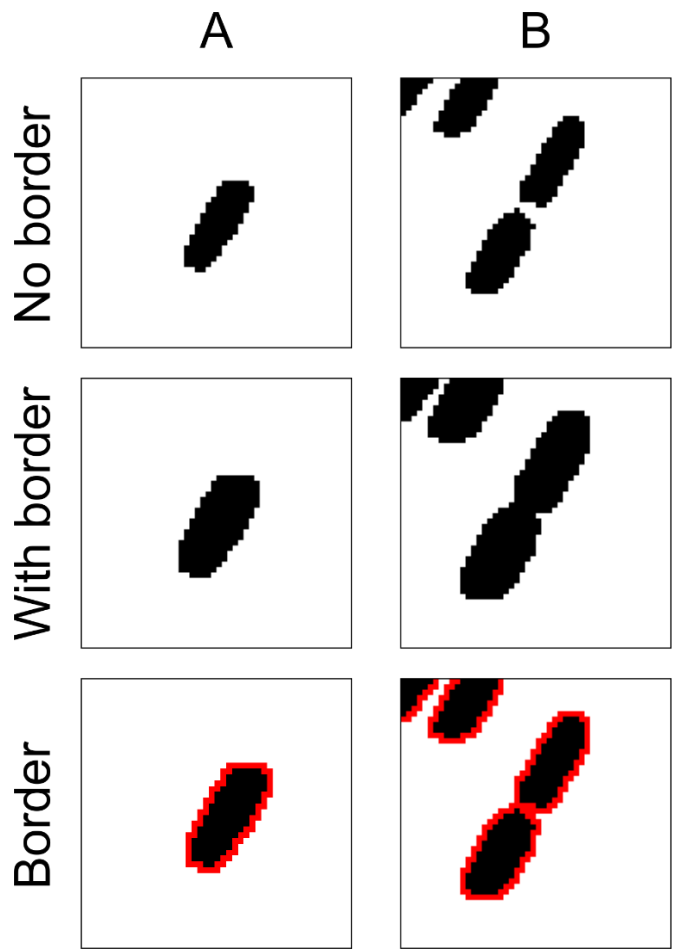

**Figure 7.** Visual comparison of the semantic to instance algorithm and the traditional semantic segmentation results.

Table 4 lists the results for the same testing areas but considering the Mask-RCNN algorithm. Region algorithms rely on some procedures to enhance the classification of small objects. The results point that using the Mask-RCNN with scaling the input image improves the results significantly. However, pixel metrics results are still far from the results using semantic segmentation architectures.

Table 4. IoU results for the Detectron2 results in the BSB vehicle dataset.

| Train # | Test Area 1 | Test Area 2 | Test Area 3 | Test Set |
|---|---|---|---|---|
| **With scale** | 80.77 | 70.09 | 59.21 | 67.65 |
| **No scale** | 74.76 | 59.54 | 49.29 | 63.34 |

*4.2.2 Per Object Metrics*

Table 5 lists the per object metrics (Correct predictions, partial predictions, false negatives, and false positives) on the three separate testing areas (Test Area 1, Test Area 2, and Test Area 3), considering the best model (containing all training samples). Test area 1 classified all objects, showing that vehicles without shadows, occlusion, and crowded areas have very high precision. On the other hand, Test Area 3, with many shadow areas and occlusion, had the highest incidence of errors, with 21 false negatives and 25 false positives. Considering that there were 430 correct predictions, the accuracy was still greater than 90%.

Table 5. Per object metrics for our proposed method.

| | Test Area 1 | Test Area 2 | Test Area 3 |
|---|---|---|---|
| Correct Predictions | 89 | 395 | 430 |
| Partial Predictions | 1 | 1 | 9 |
| False Negatives | 0 | 5 | 21 |
| False Positives | 1 | 9 | 25 |

**4.2 Semantic to Instance Segmentation Results**

Figure 8 shows three zoomed areas considering the traditional semantic segmentation method (first two rows) and our proposed box-free instance segmentation method. Both figures consider the same model. The first row (Figures 8A, 8B, and 8C) shows in yellow the merged cars, considering many vehicles in the same polygon, while the green cars were already independent even without our method. The second row (Figures 8A1, 8B1, and 8C1) shows the outlines of the polygons.

The third and fourth rows (Figures 8A2, 8B2, 8C2, 8A3, 8B3, and 8C3) show our proposed method considering the expanding border algorithm and separation into instance predictions. The fourth row shows cars in which each independent vector is represented by a different color, demonstrating that the method is efficient for separating vehicles in a precise pixel classification. Besides, interpreting these results gets much more straightforward, estimating the sizes of the vehicles and more accurate counting.

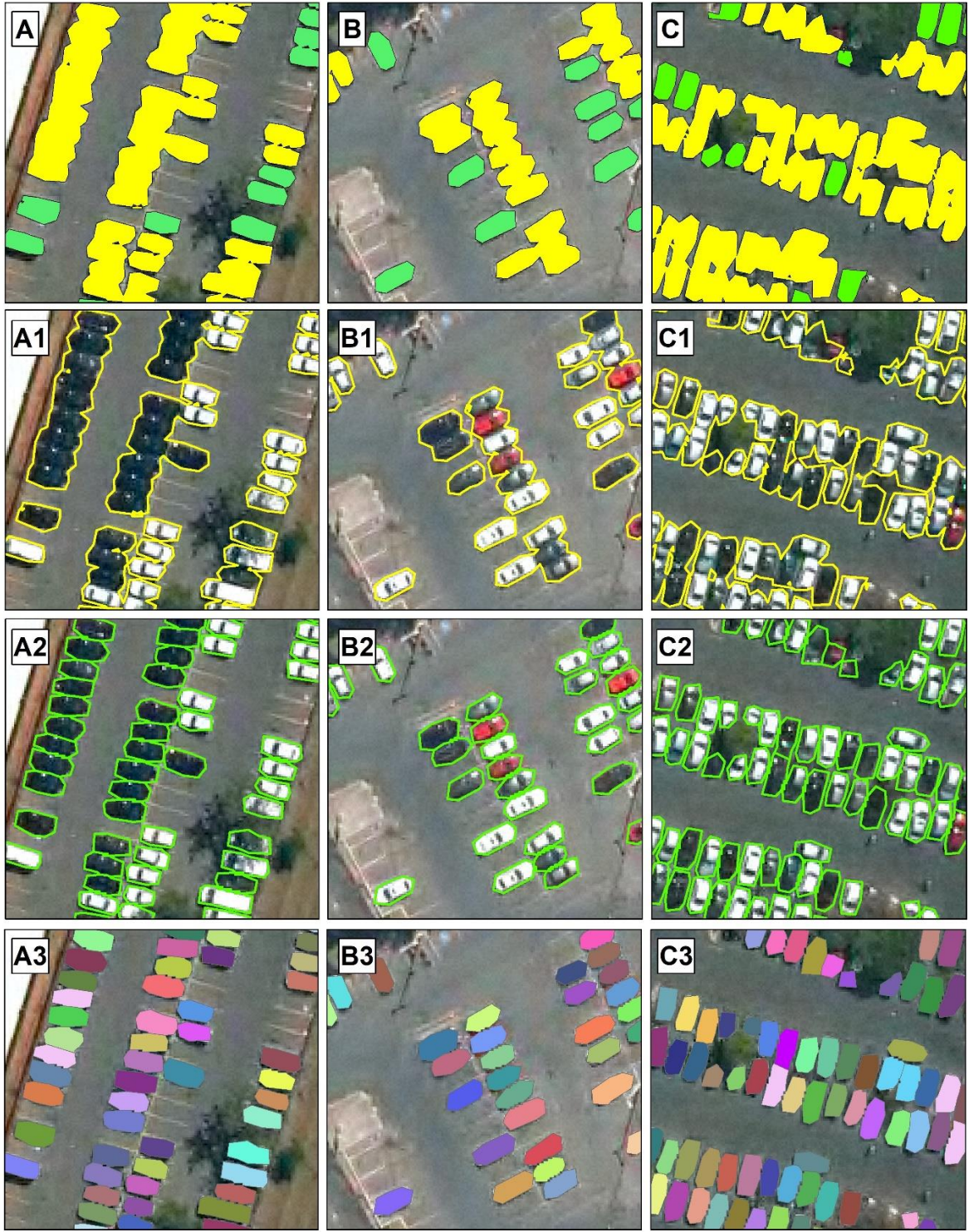

**Figure 8.** Visual comparison of the traditional semantic segmentation results without using the border procedure (first two rows), and the proposed method (last two rows).

## 4.3 Error Analysis

Even though the results were very accurate, some regions present limitations. The training procedure used many look-alikes features to train a better model. However, the number of look-alikes in a city is extensive, introducing some mistakes (Figure 9B, C, E, and F). Some crowded areas may raise some errors by joining two cars (Fig. 9A and D).

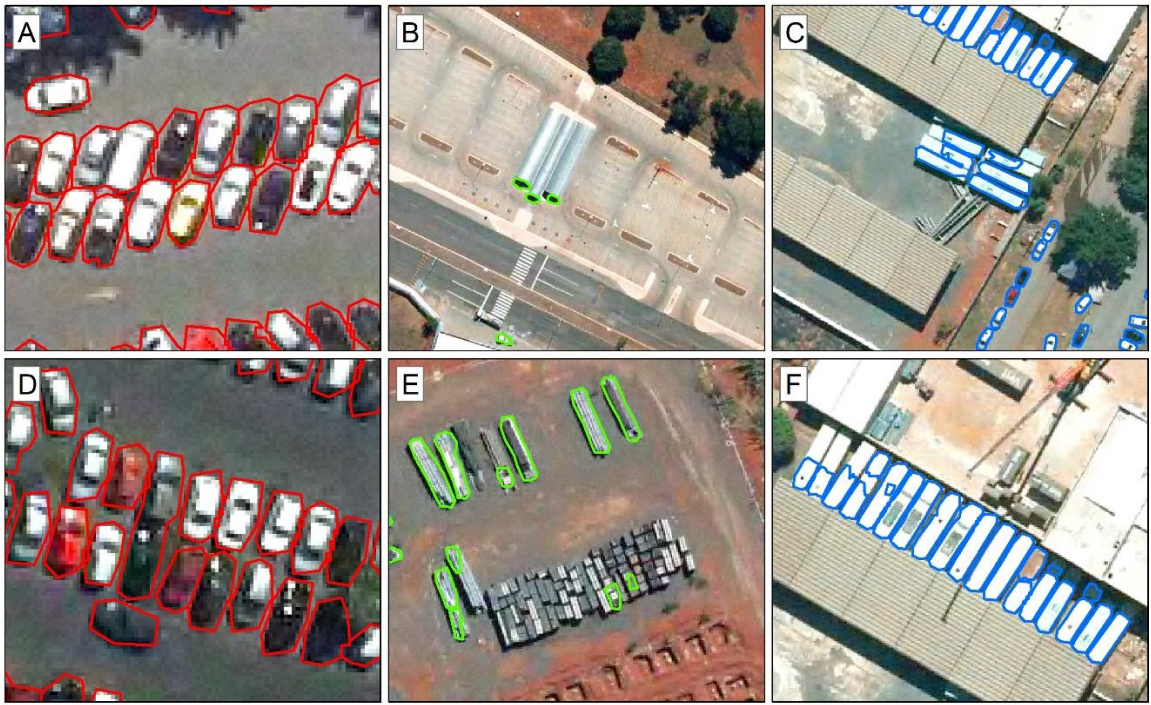

**Figure 9.** Errors in the classification procedure, and errors present from the conversion from polygon to raster.

## 4.4 Final City-Scale Classification

The final city classification presented much fewer errors when compared to the first training. However, some errors were still present, as shown in the previous section. Figure 10 shows the final classified image with a manual correction using two GIS specialists. The dataset is publicly available with more than 120 thousand vehicles (car, bus, truck, and boat).

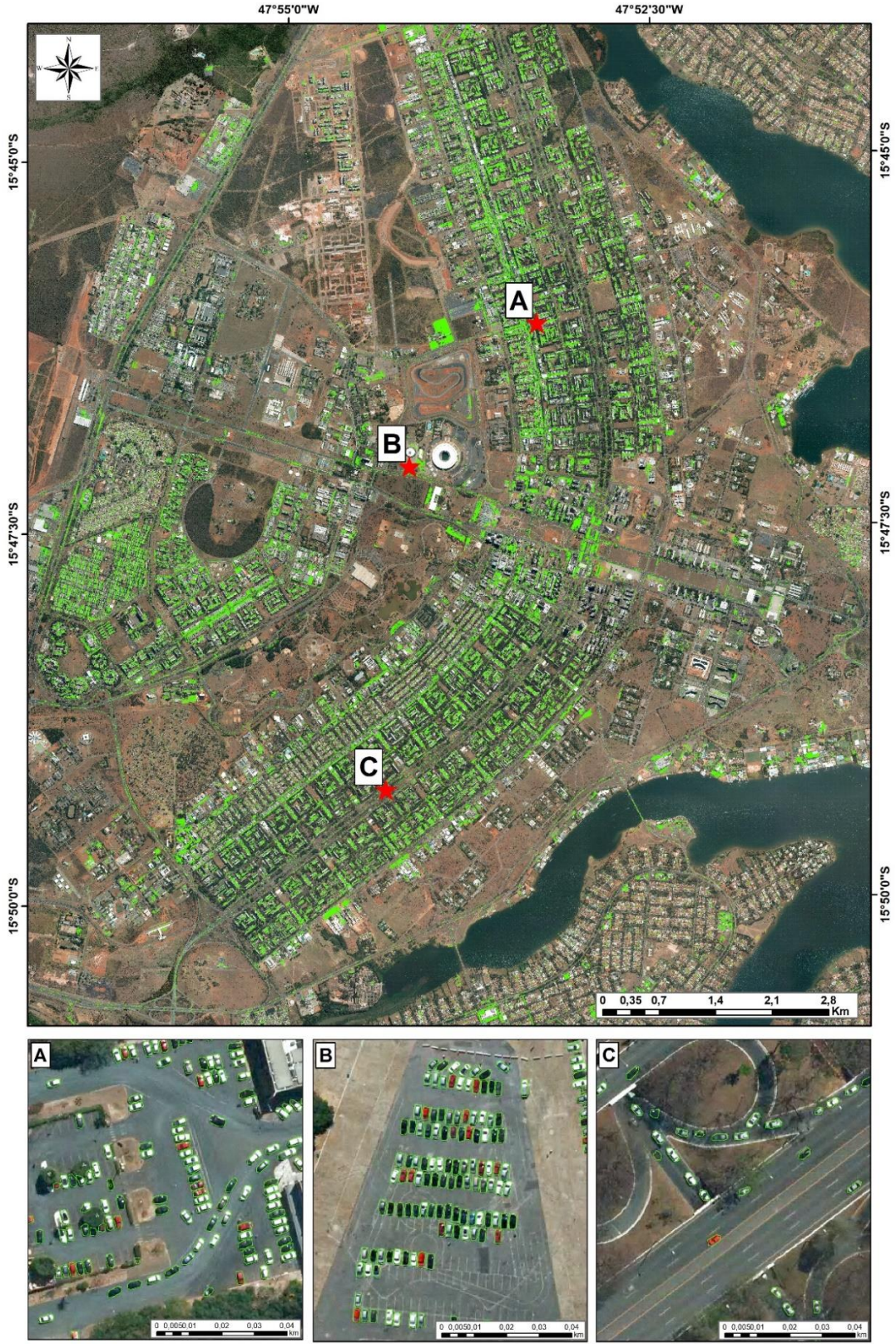

**Figure 10.** Final image classification with three zoomed areas A, B, and C.

## 4. DISCUSSION

The present research presents contributions on a dataset generation, semi supervised approach using GIS data, and a box-free instance segmentation.

### 4.3 Integration with GIS systems

To the best of our knowledge, this is the first research to use semi-supervised iterative learning with GIS platform integration. We created a tool to generate the DL samples with corresponding ground truth data for semantic (PNG mask) and instance segmentation (COCO annotation format) to extract the best out of this method. A significant advantage of this method is understanding the misclassifications zones at each iteration, enabling choosing appropriate areas to continue the training with a substantial decrease in the laborious work. Besides, generating training samples from point shapefiles allows a dataset augmentation by selecting points in strategic regions, enabling the acquisition of many samples in a limited space. This iterative approach stays in hand with Koga et al. (2018) that supply the algorithm with hard examples (e.g., look-alikes). Our method allows obtaining the exact points in which the algorithm confuses with hard examples, being able to supply those mistaken areas back to training, rapidly improving results.

Moreover, the shapefile data is easy to manipulate, correct polygons, generate borders, change classes, among others, reducing problems such as publicly available data with many errors in the ground truth data. Another great benefit is for end-users since the visualization of the data in those GIS platforms has many facilities, such as counting, choosing a specific area for analysis, getting the average size of the objects. Therefore, DL and GIS systems may work as allies for generating better predictions in less time.

### 4.2 Instance segmentation

Instance segmentation models are usually region-based proposal methods (e.g., Faster-RCNN) with a segmentation branch like the Mask-RCNN. Those models are widely used in traditional imagery, in which small objects present (area $<32^2$ pixels) presents metrics lower than medium and large objects. A common way to address the small object problem is the usage of scaling. The present research compared two situations using the Mask-RCNN (with image scaling and without image scaling). Using image scale augmentation provided the best results. However, the pixel-wise results show an inferior performance compared to the U-net with the Efficient-net-B7 backbone (67% against 82%). The generation of high-quality maps should consider distinct polygons for each object and present a good pixel-wise accuracy, demonstrating that our proposed method is very suitable for this task.

Thus, we proposed a box-free instance segmentation method using semantic segmentation models (U-net with Efficient-net-B7) with object separation by turning the interiors of the borders into distinct polygons and restoring the original object size. The usage of borders to separate instances was already presented by Mou and Zhu (2018). Even though the method is very interesting and effective, it lacks to restore the object to its original size. Our procedure uses a straightforward and fast vectorized approach using the NumPy library to recover the 1-pixel at the borders of each object.

One interesting result from our study is that the prediction not considering the borders is significantly lower than the prediction using borders, in more than 15% IoU. Besides, when increasing the vehicle dataset using iterative learning, the IoU metric considering the border rises and the IoU metric not considering the borders stays similar. This result shows us that not considering the borders for the analysis may mislead our interpretation, mainly because the error might be compensated. Since the cars present a 20x10-pixel dimension (in our dataset), a perfect prediction (100% IoU) would represent 72% IoU without considering the borders. These results show that augmenting the instance prediction borders by 1 pixel to restore the original object size is crucial for a pixel analysis, especially to understand when to stop the training procedure. In our case, the IoU from the fourth train to the fifth train only increased slightly, which would not be perceptible if considering the IoU without the borders.

Moreover, the accuracy results from our study suggest a good pixel-wise classification compared to other instance segmentation studies. Tayara et al. (2018) proposed an instance segmentation approach using points to represent each car with a Gaussian elliptical shape. Even though the method presents a good application for counting, the segmentation mask for each vehicle is not very accurate; standard elliptical shapes differ from the car shapes. Therefore, the usage of borders in each polygon is very efficient for separating different instances and maintain its shape. However, this procedure is not necessary when considering objects that do not appear crowded. For example, in tasks like swimming pool detection, the swimming pools are never adjacent since creating polygons from the segmentation masks will already present different instances, just like the cars that are not close to each other.

The generation of large areas predictions using DL is another very important topic that may be improved. Previous works show that using sliding windows with low step values corrects border errors at the frame edges, enhancing results (Audebert et al., 2017a; Costa et al., 2021; da Costa et al., 2021; de Albuquerque et al., 2020). It takes about one hour to classify our entire study area (57,856 x 42,496-pixel dimensions) using a 128-pixel stride. Future studies may evaluate the usage of parallel computing to accelerate this process.

**4.3 Vehicle Dataset**

A promising trend in Artificial Intelligence considers data-centric approaches, which consists of leveraging the data quality. In the present research, we aimed for a precise pixel-wise classification maintaining different instances for each individual object, being very relevant for vehicle studies since most vehicle datasets aim to use object detection models (only bounding boxes) (Azimi et al., 2021; Drouyer, 2020; Lin et al., 2020; Zeng et al., 2021). Some multi-class datasets also include vehicles (Xia et al., 2018; Zamir et al., 2019). The iSAID dataset only comprises vehicles, for instance, segmentation tasks, with COCO annotation format annotations. Although object detection is very promising for counting vehicles, it requires adjustments (e.g., bounding box orientation) to obtain precise information (e.g., size), making the labeling procedure more complex. Moreover, to obtain pixel information about the cars to generate a map, it is crucial to get the boundaries of each object. Our proposed method can obtain pixel-wise instance-level predictions with the same information required for a traditional semantic segmentation model, a box-free method. Furthermore, our proposed dataset stores polygonal data, facilitating additional adjustments such as dividing into more classes or refining labeled data.

Most vehicle studies use images with resolutions better than 20 cm. VAID (Lin et al., 2020) and VEDAI have the highest resolution (12.5cm) among the data sets. Our dataset has a pixel resolution of 0.24 meters, and the proposed method distinguished different instances, even at nearly twice the resolution of most datasets. The limitation of our dataset is that, for example, some distinguish sedans, which would be very difficult in our data. Therefore, our approach tends to increase efficiency with a better resolution and more suitable for separating into more classes (e.g., sedans, bus).

The resolution in this research presents information very close to WorldView3 satellite imagery. Future research may consider the usage of more spectral bands in satellite data to enhance predictions. Besides, the results on our data are much better in situations without shadows and occlusion. For the generation of aerial imagery datasets, the researchers should consider training and evaluating the data in specific day periods with fewer shadows.

## 5. CONCLUSION

The present research presented three contributions: (a) a box-free instance segmentation method, (b) a semi-supervised iterative approach to generate a high-quality dataset, and (c) the BSB vehicle dataset. The proposed DL method shows better results when compared to the Mask-RCNN architecture with a pixel-wise IoU difference greater than 12%. We show that it is crucial to consider the borders for evaluating the pixel-wise mask, being very relevant to the proposed method to restore the objects' original size. The semi-supervised iterative approach stabilized results in the fifth iteration, with a total of 1066 DL samples of 256x256 spatial dimensions. Our DL tool is a promising approach to generate datasets since it enables us to tackle strategic areas by inserting a point shapefile, significantly reducing laborious works. Finally, the BSB Vehicle Dataset was refined by two specialists containing more than 120 thousand unique vehicle polygons that are easily manipulative to other tasks.

**Data Availability Statement**

The dataset will be publicly available upon the publication of this paper. Meanwhile, the dataset may be shared with other researchers upon reasonable request to the corresponding author.

**Declaration of Competing Interest**

The authors declare that they have no known competing financial interests or personal relationships that could have appeared to influence the work reported in this paper.

**Acknowledgements**

This work was carried out with the support of the Coordination for the Improvement of Higher Education Personnel - Brazil (CAPES) - Financing Code 001. The authors are grateful for financial support from CNPq fellowship (Osmar Abílio de Carvalho Júnior, Renato Fontes Guimarães, and Roberto Arnaldo Trancoso Gomes). Special thanks are given to the research group of the Laboratory of Spatial Information System of the University of Brasilia for technical support.

**References**

Ammar, A., Koubaa, A., Ahmed, M., Saad, A., Benjdira, B., 2021. Vehicle detection from aerial images using deep learning: A comparative study. Electron. 10, 1–31. https://doi.org/10.3390/electronics10070820

Ammour, N., Alhichri, H., Bazi, Y., Benjdira, B., Alajlan, N., Zuair, M., 2017. Deep Learning Approach for Car Detection in UAV Imagery. Remote Sens. 9, 312. https://doi.org/10.3390/rs9040312

Audebert, N., Boulch, A., Randrianarivo, H., Le Saux, B., Ferecatu, M., Lefevre, S., Marlet, R., 2017a. Deep learning for urban remote sensing. 2017 Jt. Urban Remote Sens. Event, JURSE 2017. https://doi.org/10.1109/JURSE.2017.7924536

Audebert, N., Le Saux, B., Lefèvre, S., 2017b. Segment-before-Detect: Vehicle Detection and Classification through Semantic Segmentation of Aerial Images. Remote Sens. 9, 368. https://doi.org/10.3390/rs9040368

Azimi, S.M., Bahmanyar, R., Henry, C., Kurz, F., 2021. EAGLE: Large-Scale Vehicle Detection Dataset in Real-World Scenarios using Aerial Imagery, in: 2020 25th International Conference on Pattern Recognition (ICPR). IEEE, pp. 6920–6927. https://doi.org/10.1109/ICPR48806.2021.9412353

Bashir, S.M.A., Wang, Y., 2021. Small Object Detection in Remote Sensing Images with Residual Feature Aggregation-Based Super-Resolution and Object Detector Network. Remote Sens. 13, 1854. https://doi.org/10.3390/rs13091854

Benjdira, B., Khursheed, T., Koubaa, A., Ammar, A., Ouni, K., 2019. Car Detection using Unmanned Aerial Vehicles: Comparison between Faster R-CNN and YOLOv3, in: 2019 1st International Conference on Unmanned Vehicle Systems-Oman (UVS). IEEE, Muscat, Oman, pp. 1–6. https://doi.org/10.1109/UVS.2019.8658300

Cao, L., Jiang, Q., Cheng, M., Wang, C., 2016. Robust vehicle detection by combining deep features with exemplar classification. Neurocomputing 215, 225–231. https://doi.org/10.1016/j.neucom.2016.03.094

Cao, L., Luo, F., Chen, L., Sheng, Y., Wang, H., Wang, C., Ji, R., 2017. Weakly supervised vehicle detection in satellite images via multi-instance discriminative learning. Pattern Recognit. 64, 417–424. https://doi.org/10.1016/j.patcog.2016.10.033

Cao, X., Wu, C., Lan, J., Yan, P., Li, X., 2011. Vehicle detection and motion analysis in low-altitude airborne video under urban environment. IEEE Trans. Circuits Syst. Video Technol. 21, 1522–1533. https://doi.org/10.1109/TCSVT.2011.2162274

Carvalho, O.L.F. de, de Carvalho Júnior, O.A.A., Albuquerque, A.O. de, Bem, P.P. de,

Silva, C.R., Ferreira, P.H.G., Moura, R. dos S. de, Gomes, R.A.T., Guimarães, R.F., Borges, D.L.D.L., 2021. Instance Segmentation for Large, Multi-Channel Remote Sensing Imagery Using Mask-RCNN and a Mosaicking Approach. Remote Sens. 13, 39. https://doi.org/10.3390/rs13010039

Chen, C., Zhong, J., Tan, Y., 2019. Multiple-Oriented and Small Object Detection with Convolutional Neural Networks for Aerial Image. Remote Sens. 11, 2176. https://doi.org/10.3390/rs11182176

Chen, X., Xiang, S., Liu, C.L., Pan, C.H., 2014. Vehicle detection in satellite images by hybrid deep convolutional neural networks. IEEE Geosci. Remote Sens. Lett. 11, 1797–1801. https://doi.org/10.1109/LGRS.2014.2309695

Chen, Z., Wang, C., Wen, C., Teng, X., Chen, Y., Guan, H., Luo, H., Cao, L., Li, J., 2016. Vehicle detection in high-resolution aerial images via sparse representation and superpixels. IEEE Trans. Geosci. Remote Sens. 54, 103–116. https://doi.org/10.1109/TGRS.2015.2451002

Cheng, H.Y., Weng, C.C., Chen, Y.Y., 2012. Vehicle detection in aerial surveillance using dynamic bayesian networks. IEEE Trans. Image Process. 21, 2152–2159. https://doi.org/10.1109/TIP.2011.2172798

Costa, M.V.C.V. da, Carvalho, O.L.F. de, Orlandi, A.G., Hirata, I., Albuquerque, A.O. De, Silva, F.V. e, Guimarães, R.F., Gomes, R.A.T., Júnior, O.A. de C., 2021. Remote Sensing for Monitoring Photovoltaic Solar Plants in Brazil Using Deep Semantic Segmentation. Energies 14, 2960. https://doi.org/10.3390/en14102960

da Costa, L.B., de Carvalho, O.L.F., de Albuquerque, A.O., Gomes, R.A.T., Guimarães, R.F., de Carvalho Júnior, O.A., 2021. Deep semantic segmentation for detecting eucalyptus planted forests in the Brazilian territory using sentinel-2 imagery. Geocarto Int. 0, 1–13. https://doi.org/10.1080/10106049.2021.1943009

Dalal, N., Triggs, B., 2005. Histograms of Oriented Gradients for Human Detection, in: 2005 IEEE Computer Society Conference on Computer Vision and Pattern Recognition (CVPR'05). IEEE, San Diego, CA, USA, pp. 886–893. https://doi.org/10.1109/CVPR.2005.177

de Albuquerque, A.O., de Carvalho Júnior, O.A., Carvalho, O.L.F. de, de Bem, P.P., Ferreira, P.H.G., de Moura, R. dos S., Silva, C.R., Trancoso Gomes, R.A., Fontes Guimarães, R., 2020. Deep Semantic Segmentation of Center Pivot Irrigation Systems from Remotely Sensed Data. Remote Sens. 12, 2159. https://doi.org/10.3390/rs12132159

Deng, Z., Sun, H., Zhou, S., Zhao, J., Zou, H., 2017. Toward Fast and Accurate Vehicle Detection in Aerial Images Using Coupled Region-Based Convolutional Neural Networks. IEEE J. Sel. Top. Appl. Earth Obs. Remote Sens. 10, 3652–3664. https://doi.org/10.1109/JSTARS.2017.2694890

Drouyer, S., 2020. VehSat: a Large-Scale Dataset for Vehicle Detection in Satellite Images, in: IGARSS 2020 - 2020 IEEE International Geoscience and Remote Sensing Symposium. IEEE, pp. 268–271. https://doi.org/10.1109/IGARSS39084.2020.9323289

Eikvil, L., Aurdal, L., Koren, H., 2009. Classification-based vehicle detection in high-resolution satellite images. ISPRS J. Photogramm. Remote Sens. 64, 65–72. https://doi.org/10.1016/j.isprsjprs.2008.09.005

Fachrie, M., 2020. A Simple Vehicle Counting System Using Deep Learning with YOLOv3 Model. J. RESTI (Rekayasa Sist. dan Teknol. Informasi) 4, 462–468.

https://doi.org/10.29207/resti.v4i3.1871

Feng, D., Haase-Schutz, C., Rosenbaum, L., Hertlein, H., Glaser, C., Timm, F., Wiesbeck, W., Dietmayer, K., 2021. Deep Multi-Modal Object Detection and Semantic Segmentation for Autonomous Driving: Datasets, Methods, and Challenges. IEEE Trans. Intell. Transp. Syst. 22, 1341–1360. https://doi.org/10.1109/TITS.2020.2972974

Gao, Z., Ji, H., Mei, T., Ramesh, B., Liu, X., 2019. EOVNet: Earth-Observation Image-Based Vehicle Detection Network. IEEE J. Sel. Top. Appl. Earth Obs. Remote Sens. 12, 3552–3561. https://doi.org/10.1109/JSTARS.2019.2933501

Girshick, R., 2015. Fast R-CNN, in: 2015 IEEE International Conference on Computer Vision (ICCV). IEEE, Santiago, Chile, pp. 1440–1448. https://doi.org/10.1109/ICCV.2015.169

Girshick, R., Donahue, J., Darrell, T., Malik, J., 2016. Region-Based Convolutional Networks for Accurate Object Detection and Segmentation. IEEE Trans. Pattern Anal. Mach. Intell. 38, 142–158. https://doi.org/10.1109/TPAMI.2015.2437384

Girshick, R., Donahue, J., Darrell, T., Malik, J., 2014. Rich Feature Hierarchies for Accurate Object Detection and Semantic Segmentation, in: 2014 IEEE Conference on Computer Vision and Pattern Recognition. IEEE, Columbus, OH, USA, pp. 580–587. https://doi.org/10.1109/CVPR.2014.81

Gleason, J., Nefian, A. V., Bouyssounousse, X., Fong, T., Bebis, G., 2011. Vehicle detection from aerial imagery, in: 2011 IEEE International Conference on Robotics and Automation. IEEE, pp. 2065–2070. https://doi.org/10.1109/ICRA.2011.5979853

Grabner, H., Nguyen, T.T., Gruber, B., Bischof, H., 2008. On-line boosting-based car detection from aerial images. ISPRS J. Photogramm. Remote Sens. 63, 382–396. https://doi.org/10.1016/j.isprsjprs.2007.10.005

Guo, Y., Liu, Y., Georgiou, T., Lew, M.S., 2018. A review of semantic segmentation using deep neural networks. Int. J. Multimed. Inf. Retr. 7, 87–93. https://doi.org/10.1007/s13735-017-0141-z

Guo, Y., Xu, Y., Li, S., 2020. Dense construction vehicle detection based on orientation-aware feature fusion convolutional neural network. Autom. Constr. 112, 103124. https://doi.org/10.1016/j.autcon.2020.103124

Hafiz, A.M., Bhat, G.M., 2020. A survey on instance segmentation: state of the art. Int. J. Multimed. Inf. Retr. 9, 171–189. https://doi.org/10.1007/s13735-020-00195-x

Ham, S.W., Park, H.C., Kim, E.J., Kho, S.Y., Kim, D.K., 2020. Investigating the influential factors for practical application of multi-class vehicle detection for images from unmanned aerial vehicle using deep learning models. Transp. Res. Rec. 2674, 553–567. https://doi.org/10.1177/0361198120954187

Han, J., Zhang, D., Cheng, G., Guo, L., Ren, J., 2015. Object Detection in Optical Remote Sensing Images Based on Weakly Supervised Learning and High-Level Feature Learning. IEEE Trans. Geosci. Remote Sens. 53, 3325–3337. https://doi.org/10.1109/TGRS.2014.2374218

He, K., Gkioxari, G., Dollar, P., Girshick, R., 2020. Mask R-CNN. IEEE Trans. Pattern Anal. Mach. Intell. 42, 386–397. https://doi.org/10.1109/TPAMI.2018.2844175

Hinz, S., 2003. Detection and counting of cars in aerial images, in: Proceedings 2003 International Conference on Image Processing (Cat. No.03CH37429). IEEE, Barcelona, Spain, pp. III-997–1000. https://doi.org/10.1109/ICIP.2003.1247415

Holt, A.C., Seto, E.Y.W., Rivard, T., Gong, P., 2009. Object-based detection and classification of Vehicles from high-resolution aerial photography. Photogramm. Eng. Remote Sensing 75, 871–880. https://doi.org/10.14358/PERS.75.7.871

Janai, J., Güney, F., Behl, A., Geiger, A., 2020. Computer Vision for Autonomous Vehicles: Problems, Datasets and State of the Art. Found. Trends® Comput. Graph. Vis. 12, 1–308. https://doi.org/10.1561/0600000079

Javadi, S., Dahl, M., Pettersson, M.I., 2021. Vehicle Detection in Aerial Images Based on 3D Depth Maps and Deep Neural Networks. IEEE Access 9, 8381–8391. https://doi.org/10.1109/ACCESS.2021.3049741

Ji, H., Gao, Z., Mei, T., Li, Y., 2019. Improved Faster R-CNN With Multiscale Feature Fusion and Homography Augmentation for Vehicle Detection in Remote Sensing Images. IEEE Geosci. Remote Sens. Lett. 16, 1–5. https://doi.org/10.1109/LGRS.2019.2909541

Jiang, S., Yao, W., Wong, M.S., Li, G., Hong, Z., Kuc, T.Y., Tong, X., 2020. An Optimized Deep Neural Network Detecting Small and Narrow Rectangular Objects in Google Earth Images. IEEE J. Sel. Top. Appl. Earth Obs. Remote Sens. 13, 1068–1081. https://doi.org/10.1109/JSTARS.2020.2975606

Kembhavi, A., Harwood, D., Davis, L.S., 2011. Vehicle detection using partial least squares. IEEE Trans. Pattern Anal. Mach. Intell. 33, 1250–1265. https://doi.org/10.1109/TPAMI.2010.182

Koga, Y., Miyazaki, H., Shibasaki, R., 2018. A CNN-based method of vehicle detection from aerial images using hard example mining. Remote Sens. 10. https://doi.org/10.3390/rs10010124

Krizhevsky, A., Sutskever, I., Hinton, G.E., 2017. ImageNet classification with deep convolutional neural networks. Commun. ACM 60, 84–90. https://doi.org/10.1145/3065386

Leberl, F., Bischof, H., Grabner, H., Kluckner, S., 2007. Recognizing cars in aerial imagery to improve orthophotos. GIS Proc. ACM Int. Symp. Adv. Geogr. Inf. Syst. 2–10. https://doi.org/10.1145/1341012.1341015

Leitloff, J., Rosenbaum, D., Kurz, F., Meynberg, O., Reinartz, P., 2014. An operational system for estimating road traffic information from aerial images. Remote Sens. 6, 11315–11341. https://doi.org/10.3390/rs61111315

Li, Q., Mou, L., Xu, Q., Zhang, Y., Zhu, X.X., 2019. R3-Net: A Deep Network for Multioriented Vehicle Detection in Aerial Images and Videos. IEEE Trans. Geosci. Remote Sens. 57, 5028–5042. https://doi.org/10.1109/TGRS.2019.2895362

Li, Xianghui, Li, Xinde, Pan, H., 2020. Multi-Scale Vehicle Detection in High-Resolution Aerial Images with Context Information. IEEE Access 8, 208643–208657. https://doi.org/10.1109/ACCESS.2020.3036075

Liang, P., Teodoro, G., Ling, H., Blasch, E., Chen, G., Bai, L., 2012. Multiple kernel learning for vehicle detection in wide area motion imagery, in: 15th International Conference on Information Fusion, FUSION 2012. IEEE, Singapore, pp. 1629–1636.

Lin, H.-Y., Tu, K.-C., Li, C.-Y., 2020. VAID: An Aerial Image Dataset for Vehicle Detection and Classification. IEEE Access 8, 212209–212219. https://doi.org/10.1109/ACCESS.2020.3040290

Lin, T.-Y., Maire, M., Belongie, S., Hays, J., Perona, P., Ramanan, D., Dollár, P., Zitnick, C.L., 2014. Microsoft COCO: Common Objects in Context, in: Fleet, D., Tomas, P., Schiele, B., Tuytelaars, T. (Eds.), Computer Vision – ECCV 2014. Lecture Notes in Computer Science, Vol 8693. Springer Cham, Zurich, Switzerland, pp. 740–755.

https://doi.org/10.1007/978-3-319-10602-1_48

Liu, C., Ding, Y., Zhu, M., Xiu, J., Li, M., Li, Q., 2019. Vehicle Detection in Aerial Images Using a Fast Oriented Region Search and the Vector of Locally Aggregated Descriptors. Sensors 19, 3294. https://doi.org/10.3390/s19153294

Liu, K., Mattyus, G., 2015. Fast Multiclass Vehicle Detection on Aerial Images. IEEE Geosci. Remote Sens. Lett. 12, 1938–1942. https://doi.org/10.1109/LGRS.2015.2439517

Liu, W., Anguelov, D., Erhan, D., Szegedy, C., Reed, S., Fu, C.-Y., Berg, A.C., 2016. SSD: Single Shot MultiBox Detector, in: Computer Vision – ECCV 2016. Lecture Notes in Computer Science. Springer, Cham, pp. 21–37. https://doi.org/10.1007/978-3-319-46448-0_2

Liu, X., Yang, T., Li, J., 2018. Real-time ground vehicle detection in aerial infrared imagery based on convolutional neural network. Electron. 7, 1–19. https://doi.org/10.3390/electronics7060078

Madhogaria, S., Baggenstoss, P., Schikora, M., Koch, W., Cremers, D., 2015. Car detection by fusion of HOG and causal MRF. IEEE Trans. Aerosp. Electron. Syst. 51, 575–590. https://doi.org/10.1109/TAES.2014.120141

Mandal, M., Shah, M., Meena, P., Devi, S., Vipparthi, S.K., 2020. AVDNet: A Small-Sized Vehicle Detection Network for Aerial Visual Data. IEEE Geosci. Remote Sens. Lett. 17, 494–498. https://doi.org/10.1109/LGRS.2019.2923564

Moranduzzo, T., Melgani, F., 2014a. Automatic car counting method for unmanned aerial vehicle images. IEEE Trans. Geosci. Remote Sens. 52, 1635–1647. https://doi.org/10.1109/TGRS.2013.2253108

Moranduzzo, T., Melgani, F., 2014b. Detecting cars in UAV images with a catalog-based approach. IEEE Trans. Geosci. Remote Sens. 52, 6356–6367. https://doi.org/10.1109/TGRS.2013.2296351

Mou, L., Zhu, X.X., 2018. Vehicle Instance Segmentation From Aerial Image and Video Using a Multitask Learning Residual Fully Convolutional Network. IEEE Trans. Geosci. Remote Sens. 56, 6699–6711. https://doi.org/10.1109/TGRS.2018.2841808

Nguyen, T.T., Grabner, H., Bischof, H., Gruber, B., 2007. On-line boosting for car detection from aerial images. 2007 IEEE Int. Conf. Res. Innov. Vis. Futur. RIVF 2007 87–95. https://doi.org/10.1109/RIVF.2007.369140

Ophoff, T., Puttemans, S., Kalogirou, V., Robin, J.-P., Goedemé, T., 2020. Vehicle and Vessel Detection on Satellite Imagery: A Comparative Study on Single-Shot Detectors. Remote Sens. 12, 1217. https://doi.org/10.3390/rs12071217

Qu, S., Wang, Y., Meng, G., Pan, C., 2016. Vehicle Detection in Satellite Images by Incorporating Objectness and Convolutional Neural Network. J. Ind. Intell. Inf. 4, 158–162. https://doi.org/10.18178/jiii.4.2.158-162

Razakarivony, S., Jurie, F., 2016. Vehicle detection in aerial imagery: A small target detection benchmark. J. Vis. Commun. Image Represent. 34, 187–203. https://doi.org/10.1016/j.jvcir.2015.11.002

Redmon, J., Divvala, S., Girshick, R., Farhadi, A., 2016. You Only Look Once: Unified, Real-Time Object Detection, in: 2016 IEEE Conference on Computer Vision and Pattern Recognition (CVPR). IEEE, Las Vegas, NV, USA, pp. 779–788. https://doi.org/10.1109/CVPR.2016.91

Reksten, J.H., Salberg, A.B., 2021. Estimating Traffic in Urban Areas from Very-High Resolution Aerial Images. Int. J. Remote Sens. 42, 865–883.

https://doi.org/10.1080/01431161.2020.1815891

Ren, S., He, K., Girshick, R., Sun, J., 2017. Faster R-CNN: Towards Real-Time Object Detection with Region Proposal Networks. IEEE Trans. Pattern Anal. Mach. Intell. 39, 1137–1149. https://doi.org/10.1109/TPAMI.2016.2577031

Ronneberger, O., Fischer, P., Brox, T., 2015. U-Net: Convolutional Networks for Biomedical Image Segmentation, in: Navab, N., Hornegger, J., Wells, W., Frangi, A. (Eds.), Lecture Notes in Computer Science (Including Subseries Lecture Notes in Artificial Intelligence and Lecture Notes in Bioinformatics). Springer, Cham, pp. 234–241. https://doi.org/10.1007/978-3-319-24574-4_28

Sakhare, K. V., Tewari, T., Vyas, V., 2020. Review of Vehicle Detection Systems in Advanced Driver Assistant Systems. Arch. Comput. Methods Eng. 27, 591–610. https://doi.org/10.1007/s11831-019-09321-3

Sevo, I., Avramovic, A., 2016. Convolutional Neural Network Based Automatic Object Detection on Aerial Images. IEEE Geosci. Remote Sens. Lett. 13, 740–744. https://doi.org/10.1109/LGRS.2016.2542358

Shao, W., Yang, W., Liu, G., Liu, J., 2012. Car detection from high-resolution aerial imagery using multiple features. Int. Geosci. Remote Sens. Symp. 4379–4382. https://doi.org/10.1109/IGARSS.2012.6350403

Sharma, G., Merry, C.J., Goel, P., McCord, M., 2006. Vehicle detection in 1-m resolution satellite and airborne imagery. Int. J. Remote Sens. 27, 779–797. https://doi.org/10.1080/01431160500238901

Shen, J., Liu, N., Sun, H., 2021. Vehicle detection in aerial images based on lightweight deep Convolutional network. IET Image Process. 15, 479–491. https://doi.org/10.1049/ipr2.12038

Shen, J., Liu, N., Sun, H., 2019. Vehicle Detection in Aerial Images Based on Hyper Feature Map in Deep Convolutional Network. KSII Trans. Internet Inf. Syst. 13, 479–491. https://doi.org/10.3837/tiis.2019.04.014

Shi, F., Zhang, Tong, Zhang, Tao, 2021. Orientation-Aware Vehicle Detection in Aerial Images via an Anchor-Free Object Detection Approach. IEEE Trans. Geosci. Remote Sens. 59, 5221–5233. https://doi.org/10.1109/TGRS.2020.3011418

Sommer, L., Schuchert, T., Beyerer, J., 2019. Comprehensive Analysis of Deep Learning-Based Vehicle Detection in Aerial Images. IEEE Trans. Circuits Syst. Video Technol. 29, 2733–2747. https://doi.org/10.1109/TCSVT.2018.2874396

Song, H., Liang, H., Li, H., Dai, Z., Yun, X., 2019. Vision-based vehicle detection and counting system using deep learning in highway scenes. Eur. Transp. Res. Rev. 11. https://doi.org/10.1186/s12544-019-0390-4

Stuparu, D.G., Ciobanu, R.I., Dobre, C., 2020. Vehicle detection in overhead satellite images using a one-stage object detection model. Sensors (Switzerland) 20, 1–18. https://doi.org/10.3390/s20226485

Tan, M., Le, Q. V., 2019. EfficientNet: Rethinking Model Scaling for Convolutional Neural Networks. arXiv.

Tan, Q., Ling, J., Hu, Jun, Qin, X., Hu, Jiping, 2020. Vehicle Detection in High Resolution Satellite Remote Sensing Images Based on Deep Learning. IEEE Access 8, 153394–153402. https://doi.org/10.1109/ACCESS.2020.3017894

Tang, T., Zhou, S., Deng, Z., Zou, H., Lei, L., 2017. Vehicle Detection in Aerial Images Based on Region Convolutional Neural Networks and Hard Negative Example Mining. Sensors 17, 336. https://doi.org/10.3390/s17020336

Tao, C., Mi, L., Li, Y., Qi, J., Xiao, Y., Zhang, J., 2019. Scene Context-Driven Vehicle Detection in High-Resolution Aerial Images. IEEE Trans. Geosci. Remote Sens. 57, 7339–7351. https://doi.org/10.1109/TGRS.2019.2912985

Tayara, H., Gil Soo, K., Chong, K.T., 2018. Vehicle Detection and Counting in High-Resolution Aerial Images Using Convolutional Regression Neural Network. IEEE Access 6, 2220–2230. https://doi.org/10.1109/ACCESS.2017.2782260

Tuermer, S., Kurz, F., Reinartz, P., Stilla, U., 2013. Airborne vehicle detection in dense urban areas using HoG features and disparity maps. IEEE J. Sel. Top. Appl. Earth Obs. Remote Sens. 6, 2327–2337. https://doi.org/10.1109/JSTARS.2013.2242846

Van Etten, A., 2018. You Only Look Twice: Rapid Multi-Scale Object Detection In Satellite Imagery.

Viola, P., Jones, M., 2001. Rapid object detection using a boosted cascade of simple features, in: Proceedings of the 2001 IEEE Computer Society Conference on Computer Vision and Pattern Recognition. CVPR 2001. IEEE Comput. Soc, Kauai, HI, USA, pp. I-511-I–518. https://doi.org/10.1109/CVPR.2001.990517

Wang, B., Gu, Y., 2020. An improved FBPN-based detection network for vehicles in aerial images. Sensors (Switzerland) 20, 1–20. https://doi.org/10.3390/s20174709

Wang, H., Yu, Y., Cai, Y., Chen, X., Chen, L., Liu, Q., 2019. A Comparative Study of State-of-the-Art Deep Learning Algorithms for Vehicle Detection. IEEE Intell. Transp. Syst. Mag. 11, 82–95. https://doi.org/10.1109/MITS.2019.2903518

Wang, J., Simeonova, S., Shahbazi, M., 2019. Orientation- and Scale-Invariant Multi-Vehicle Detection and Tracking from Unmanned Aerial Videos. Remote Sens. 11, 2155. https://doi.org/10.3390/rs11182155

Wu, Y., Kirillov, A., Massa, F., Lo, W.-Y., Girshick, R., 2019. Detectron2 [WWW Document]. URL https://github.com/facebookresearch/detectron2 (accessed 3.3.21).

Xi, X., Yu, Z., Zhan, Z., Yin, Y., Tian, C., 2019. Multi-Task Cost-Sensitive-Convolutional Neural Network for Car Detection. IEEE Access 7, 98061–98068. https://doi.org/10.1109/ACCESS.2019.2927866

Xia, G., Bai, X., Ding, J., Zhu, Z., Belongie, S., Luo, J., Datcu, M., Pelillo, M., Zhang, L., 2018. DOTA: A Large-Scale Dataset for Object Detection in Aerial Images, in: 2018 IEEE/CVF Conference on Computer Vision and Pattern Recognition. IEEE, pp. 3974–3983. https://doi.org/10.1109/CVPR.2018.00418

Xie, S., Girshick, R., Dollar, P., Tu, Z., He, K., 2017. Aggregated Residual Transformations for Deep Neural Networks, in: 2017 IEEE Conference on Computer Vision and Pattern Recognition (CVPR). IEEE, Honolulu, HI, USA, pp. 5987–5995. https://doi.org/10.1109/CVPR.2017.634

Xu, Y., Yu, G., Wang, Y., Wu, X., Ma, Y., 2017a. Car detection from low-altitude UAV imagery with the faster R-CNN. J. Adv. Transp. 2017. https://doi.org/10.1155/2017/2823617

Xu, Y., Yu, G., Wang, Y., Wu, X., Ma, Y., 2016. A Hybrid Vehicle Detection Method Based on Viola-Jones and HOG + SVM from UAV Images. Sensors 16, 1325. https://doi.org/10.3390/s16081325

Xu, Y., Yu, G., Wu, X., Wang, Y., Ma, Y., 2017b. An Enhanced Viola-Jones Vehicle Detection Method From Unmanned Aerial Vehicles Imagery. IEEE Trans. Intell. Transp. Syst. 18, 1845–1856. https://doi.org/10.1109/TITS.2016.2617202

Yakubovskiy, P., 2020. Segmentation Models Pytorch. GitHub Repos.

Yang, M.Y., Liao, W., Li, X., Cao, Y., Rosenhahn, B., 2019. Vehicle detection in aerial

images. Photogramm. Eng. Remote Sensing 85, 297–304. https://doi.org/10.14358/PERS.85.4.297

Yu, Y., Gu, T., Guan, H., Li, D., Jin, S., 2019. Vehicle Detection from High-Resolution Remote Sensing Imagery Using Convolutional Capsule Networks. IEEE Geosci. Remote Sens. Lett. 16, 1894–1898. https://doi.org/10.1109/LGRS.2019.2912582

Yu, Y., Guan, H., Ji, Z., 2015. Rotation-Invariant Object Detection in High-Resolution Satellite Imagery Using Superpixel-Based Deep Hough Forests. IEEE Geosci. Remote Sens. Lett. 12, 2183–2187. https://doi.org/10.1109/LGRS.2015.2432135

Yu, Y., Guan, H., Zai, D., Ji, Z., 2016. Rotation-and-scale-invariant airplane detection in high-resolution satellite images based on deep-Hough-forests. ISPRS J. Photogramm. Remote Sens. 112, 50–64. https://doi.org/10.1016/j.isprsjprs.2015.04.014

Zamir, S.W., Arora, A., Gupta, A., Khan, S., Sun, G., Khan, F.S., Zhu, F., Shao, L., Xia, G.S., Bai, X., 2019. iSAID: A large-scale dataset for instance segmentation in aerial images. arXiv.

Zeng, Y., Duan, Q., Chen, X., Peng, D., Mao, Y., Yang, K., 2021. UAVData: A dataset for unmanned aerial vehicle detection. Soft Comput. 25, 5385–5393. https://doi.org/10.1007/s00500-020-05537-9

Zhang, X., Zhu, X., 2019. An Efficient and Scene-Adaptive Algorithm for Vehicle Detection in Aerial Images Using an Improved YOLOv3 Framework. ISPRS Int. J. Geo-Information 8, 483. https://doi.org/10.3390/ijgi8110483

Zhao, Z.Q., Zheng, P., Xu, S.T., Wu, X., 2019. Object Detection with Deep Learning: A Review. IEEE Trans. Neural Networks Learn. Syst. 30, 3212–3232. https://doi.org/10.1109/TNNLS.2018.2876865

Zheng, Z., Zhou, G., Wang, Y., Liu, Y., Li, X., Wang, X., Jiang, L., 2013. A Novel Vehicle Detection Method With High Resolution Highway Aerial Image. IEEE J. Sel. Top. Appl. Earth Obs. Remote Sens. 6, 2338–2343. https://doi.org/10.1109/JSTARS.2013.2266131

Zhong, J., Lei, T., Yao, G., 2017. Robust vehicle detection in aerial images based on cascaded convolutional neural networks. Sensors (Switzerland) 17. https://doi.org/10.3390/s17122720

Zhou, H., Wei, L., Lim, C.P., Creighton, D., Nahavandi, S., 2018. Robust vehicle detection in aerial images using bag-of-words and orientation aware scanning. IEEE Trans. Geosci. Remote Sens. 56, 7074–7085. https://doi.org/10.1109/TGRS.2018.2848243

Zhu, J., Sun, K., Jia, S., Li, Q., Hou, X., Lin, W., Liu, B., Qiu, G., 2018. Urban Traffic Density Estimation Based on Ultrahigh-Resolution UAV Video and Deep Neural Network. IEEE J. Sel. Top. Appl. Earth Obs. Remote Sens. 11, 4968–4981. https://doi.org/10.1109/JSTARS.2018.2879368